\documentclass[letterpaper, 10 pt]{IEEEtran}
\IEEEoverridecommandlockouts

\usepackage{url}
\usepackage[x11names]{xcolor}
\usepackage[colorlinks, bookmarksopen, bookmarksnumbered, citecolor=Red4, urlcolor=Turquoise4, linkcolor=Green4]{hyperref}

\usepackage{natbib}
\usepackage{amsmath}
\usepackage{bm}
\usepackage{amssymb}
\usepackage{graphicx}
\usepackage{mathtools}
\usepackage{stmaryrd}
\usepackage{balance}
\usepackage{booktabs}
\usepackage{multirow}
\usepackage{siunitx}
\usepackage[caption=false]{subfig}

\DeclareMathOperator*{\argmax}{arg\,max}

\title{\Large \bf Data-driven control of hydraulic impact hammers under strict operational and control constraints}

\author{Francisco Leiva$^{1}$, Claudio Canales$^{1}$, Michelle Valenzuela$^{1}$ and Javier Ruiz-del-Solar$^{1}$%
\thanks{This work was supported by FONDECYT project 1251823, ANID-PIA project CIA250010, and ANID/Doctorado Nacional/2023-21232021.}%
\thanks{$^1$Advanced Mining Technology Center (AMTC) and Department of Electrical Engineering, Universidad de Chile, Tupper 2007, Santiago, Chile.}%
\thanks{\tt\small{francisco.leiva@ing.uchile.cl, claudio.canales@amtc.uchile.cl, michelle.valenzuela@amtc.uchile.cl, jruizd@ing.uchile.cl}}%
}

\begin{document}

\maketitle

\begin{abstract}

This paper presents a data-driven methodology for the control of static hydraulic impact hammers, also known as rock breakers, which are commonly used in the mining industry. The task addressed in this work is that of controlling the rock-breaker so its end-effector reaches arbitrary target poses from any given initial configuration, which is required in normal operation to place the hammer on top of rocks that need to be fractured. The proposed approach considers several constraints, such as unobserved state variables due to limited sensing and the strict requirement of using a discrete control interface at the joint level. First, the proposed methodology addresses the problem of system identification in order to obtain an approximate dynamic model of the hydraulic arm. This is done via supervised learning, using only teleoperation data. The learned dynamic model is then exploited to obtain a controller capable of reaching target end-effector poses. For policy synthesis, both reinforcement learning (RL) and model predictive control (MPC) algorithms are utilized and contrasted. As a case study, we consider the automation of a Bobcat~E10 mini-excavator arm with a hydraulic impact hammer attached as end-effector. Using this machine, both the system identification and policy synthesis stages are extensively studied in simulation and in the real world. The best RL-based policy consistently reaches target end-effector poses with position errors below 12 [cm] and pitch angle errors below 0.08 [rad] when deployed in the real world. Considering that the impact hammer has a 4 [cm] diameter chisel, this level of precision is sufficient for breaking rocks. Notably, this is accomplished by relying only on approximately 68~min of teleoperation data to train and 8~min to evaluate the dynamic model, and without performing any adjustments or fine-tuning for a successful policy Sim2Real transfer. A demonstration of policy execution in the real world can be found in \url{https://youtu.be/e-7tDhZ4ZgA}.

\end{abstract}


\section{Introduction}
\label{sec:introduction}

The increasing demand for automation comes with many challenges. This is particularly true when automating heavy-duty hydraulic machinery operating in unstructured environments. The automation of such systems is difficult due to their highly nonlinear dynamics, delayed responses, and the constraints associated with intervening manually operated machines that are already deployed in the field.

\begin{figure}
    \centering
    \includegraphics[width=\linewidth]{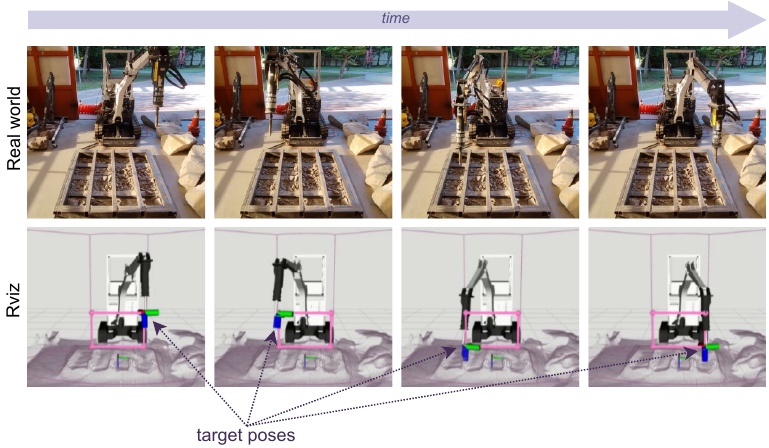}
    \caption{Depiction of the reaching task performed by impact hammers in mining, using a Bobcat E10 mini-excavator. To perform this task, the hydraulic arm of the machine has to be controlled so its end-effector reaches target poses from arbitrary initial configurations, using minimal sensing and a discrete control interface at the joint level.}
    \label{fig:reaching_depiction}
\end{figure}

This work addresses the automation of hydraulic impact hammers (also known as rock-breakers) used in mining. These machines can be described as static hydraulic arms with an hydraulic impact hammer attached as an end-effector. Operators remotely control these machines to break rocks that are too large to fall through steel grates installed on top of ore passes, allowing material to be transported to lower production levels~\citep{correa2022haptic}. Particularly, we address the problem of synthesizing a controller capable of reaching end-effector target poses (for instance, rock breaking poses) for these machines. This task, commonly known as \emph{``reaching''}, is constantly performed by human operators to place the impact hammer on top of rocks that need to be fractured, or to place it in safe areas whenever material is deposited on the steel grates by a load-haul-dump (LHD) machine. Fig.~\ref{fig:reaching_depiction} illustrates the problem of reaching arbitrary target end-effector poses using a mini-excavator whose arm has the the same kinematic structure that full-sized rock breakers, and that has an hydraulic impact hammer attached as end-effector.

The main challenges associated with the automation of hydraulic impact hammers used in mining arise from the prohibition of any physical modification on the machines (which restricts sensing), the inability to accurately capture the system dynamics (as proprietary components often conceal their internal responses), and the requirement that control must be executed through discrete commands at the joint level (since this allows standardizing a common control interface for a fleet of impact hammers). These constraints originate from industrial requirements aiming to minimize intervention on machines that are already actively operating in production. 

To achieve the automation goal given the above constraints, a two-stage data-driven methodology is adopted. In the first stage, the system's dynamics is approximated by training a parameterized function using operational data, and then, in the second stage, the obtained dynamic model is exploited to synthesize reaching controllers. The synthesis of policies is conducted by leveraging an efficient computational pipeline, in which the learned model is used to predict the real system's response to control commands. This allows generating experiences that are used to learn a controller using reinforcement learning (RL). In addition, a model predictive control (MPC) scheme is also implemented as a baseline. The obtained policies are then deployed and tested in the real world without performing any fine-tuning.

This methodology is implemented and validated to synthesize a reaching controller for a hydraulic mini-excavator. This mini-excavator is electro-hydraulically intervened to allow its automation, and in a manner that emulates the control challenges described for static rock-breakers used in mining. This results in unobserved state variables that makes classical dynamic modeling non-viable, and in a discrete control interface that makes accurately reaching target poses challenging.

With the above, the main contributions of this work are the following:

\begin{itemize}
    
    \item A practical methodology for automating hydraulic arms with unobserved variables and a discrete control interface, covering the full pipeline from system identification to controller deployment. By relying on a minimal sensing setup, the approach can be transferred to other machines with minimal adaptation. The results demonstrate that effective hydraulic control can be achieved even in conditions of low observability and with limited ground truth data, thereby streamlining the deployment of industrial automation.
    \item A policy (obtained by exploiting a learned data-driven model) that perform the reaching task in a three-dimensional space. Unlike most existing controllers, the policy operates in a discrete action space, dictated by the hardware and operational requirements of the target application (secondary rock reduction in mining using hydraulic breakers). This design choice introduces additional complexity in both training and implementation. Despite these challenges, the controller achieves high accuracy, consistently reaching target poses with a position error below $12$~[cm] and an absolute orientation error below $0.08$~[rad]. Remarkably, the policy is transferred directly from simulation to the real machine without any parameter tuning.
    
\end{itemize}

\section{Related work}
\label{sec:related_work}

The automation of industrial machinery has been extensively explored in recent years, driven by the demand for increased efficiency, safety, and productivity in complex work environments. In this domain, hydraulic machines are unique because of their power and versatility. These machines are widely used in construction, mining, forestry, and demolition, where they perform physically demanding tasks such as excavation, rock breaking, material handling, and infrastructure maintenance. Their broad range of applications has made them a focal point in the study of robotic automation. Consequently, hydraulic system automation has been an active area of research for several years, with numerous works that address the unique challenges posed by their nonlinear dynamics, delayed responses, and the need for precise control under uncertain conditions~\citep{Mattila2017ASO}.

Traditional approaches to hydraulic machine automation are predominantly model-based. Early works such as~\cite{Plummer1990RobustAC, Sohl1997ExperimentsAS, Habibi1991ComputedTorqueAV,Sirouspour2001NonlinearCO} aimed to achieve control by deriving dynamic models of the hydraulic system and applying different control techniques. For instance, in~\cite{Sohl1997ExperimentsAS}, a non-linear tracking controller based on Lyapunov theory and backstepping is presented for a hydraulic servosystem, guaranteeing exponential stability for force tracking. However, the approach relies heavily on accurate parameter identification, such as valve gains and fluid bulk modulus, and requires access to signal derivatives, which are often noisy in practice. This class of requirements are difficult to fulfill in real-world industrial settings, often leading to complicated or limited implementations, compromising effectiveness, scalability, and robustness. This challenge is common across many classical model-based control strategies.

Even today, model-based approaches remain an active area of research in hydraulic control. Modern studies continue to explore analytical control strategies, often combining model-based formulations with advanced techniques such as adaptive control, sliding mode control (SMC) and model predictive control (MPC). These methods frequently incorporate online parameter estimation or disturbance observers to improve performance under uncertainty. For example, in \cite{Bender2017ModelingAO}, an offset‑free MPC controller is implemented in a hydraulic mini-excavator, allowing precise control of the arm movement while handling disturbances. However, these methods still inherently rely on the assumed model structure, which remains a critical weakness.

In parallel, model-free control strategies remain widely used in practice, particularly proportional-integral-derivative (PID) control (e.g.,~\cite{egli2024reinforcement}), due to their simplicity and ease of implementation. Nevertheless, PID controllers require extensive tuning for each machine, and their performance tends to degrade as the system moves away from the operating point around which the gains were calibrated.

Given the difficulty of deriving accurate analytical models for hydraulic machines, data-driven approaches have long been considered a promising alternative. In \cite{Song1995NeuralAC}, it was proposed to learn the inverse dynamics of the plant (an hydraulic excavator) using operational data. More recently, this data-driven strategy has gained renewed interest, driven by advances in machine learning techniques.  Works such as \cite{Park2017OnlineLC}, \cite{lee2022precision}, \cite{Ma2024DataDrivenMN}, \cite{ma2024data} and \cite{Greiser2024FeedforwardCF} use modern machine learning techniques to model complex system behaviors from sensor and control data. While these approaches offer clear advantages, such as adaptability, reduced modeling effort, and the ability to capture non-linearities, they are not without challenges. Issues such as data quality, generalization to unseen conditions, or limited interpretability remain active areas of concern.

An alternative strategy to control hydraulic machines and learn complex tasks is reinforcement learning (RL). One of the main advantages of RL is its ability to learn control policies directly from interaction with the environment, without necessarily requiring an accurate model of the system dynamics. This is particularly appealing for hydraulic systems, which are notoriously difficult to model due to strong nonlinearities and complex internal dynamics, making classical control design a highly challenging task. Moreover, RL methods have shown promising results in simulation and even on real machines, successfully learning behaviors that are difficult to encode explicitly using conventional control techniques.

However, it is important to note that RL is not without drawbacks. The design of a reward function that effectively captures the objective of a task often requires extensive domain knowledge and entails a substantial trial-and-error process. Additionally, policies trained in simulation may fail to transfer to real-world machines due to discrepancies between simulated and real dynamics, a well known issue referred to as the sim-to-real gap. To mitigate this, it is crucial to ensure that the simulated dynamics closely approximate those of the real system. As seen in \cite{egli2020towards}, \cite{Egli2022AGA}, \cite{spinelli2024reinforcement} and \cite{spinelli2025large}, a promising way to tackle this issue is the use of a data-driven approach to approximate the plant dynamics and then to rely on this learned model to synthesize an RL-based policy.
 
It is also worth mentioning that, even if many of these approaches offer promising results, they rely strongly on the observability of the system. For example, in works such as those of ~\cite{Jud2021HEAPT}, \cite{egli2020towards}, \cite{Egli2022AGA},  and \cite{spinelli2024reinforcement}, precise measurements for the joint positions and velocities, often alongside pressures, torques, engine RPM measurements, among others, are utilized to characterize the instantaneous state of the machines being controlled.

Regarding the automation of hydraulic impact hammers (rock breakers) used in mining, prior work has reported the development of teleoperation, assisted teleoperation~\citep{correa2022haptic}, controllers that solve a particular task of the full operating cycle of the rock breaker (e.g. rock breaking in~\cite{samtani2023learning}) and complete modular automation systems~\citep{lampinen2021autonomous}. 

In~\cite{lampinen2021autonomous}, for example, motion control is achieved by learning a mapping between input control signals and the resulting velocity for each valve-actuator pair, following the approach proposed by~\cite{nurmi2017automated}. These learned models are then used to control the machine to track trajectories generated by a planner that takes into account the flow restrictions of the hammer's hydraulic unit, using the algorithm proposed by~\cite{lampinen2020flow}. In~\cite{samtani2023learning} only the rock-breaking stage of the machine's operation cycle (which starts once the end-effector is positioned over a rock) is automated via model-free RL.

Given the state of the art, this work follows the data-driven approach adopted by many recent works (e.g.,~\cite{egli2020towards,Egli2022AGA}): learning a model of the system dynamics, and then obtaining a policy by exploiting the learned model. However, in contrast to the existing literature, this work addresses the challenge of learning a dynamic model by relying on a minimal set of measured variables, namely, only joint positions, and performing the policy synthesis stage considering a discrete action space. Note that both restrictions arise from operational constraints on the target application and, while~\cite{lampinen2021autonomous} achieve good results regarding the motion control of a full-sized impact hammer, their method is not directly applicable in these settings, since their control law is designed for a continuous action space. Moreover, in contrast to works like those of~\cite{lampinen2020flow} and~\cite{Egli2022AGA}, we address the problem of reaching an arbitrary target pose from a given initial configuration, without the need for a full plan or waypoints.

\section{Problem description}

\subsection{Automation of impact hammers in underground mining}

Impact hammers, also known as rock breakers, are hydraulic arms, typically with four degrees of freedom and a hydraulic impact hammer as end-effector. These machines are widely used in underground mining, where they perform the task of fracturing rocks that cannot pass through transfer steel grates installed on top of ore passes, which allows the material to be transported to a lower production level due to gravity~\citep{correa2022haptic}.

Impact hammers used in these settings are installed adjacent to steel grates and are remotely controlled by expert human operators, who work in a safe control room, usually many kilometers away from the mine. The operators controlling the impact hammers do so by remotely actuating its electro-valves using a joystick, and by relying mainly on visual information of the environment, which is provided by CCTV cameras installed near the steel grates.

Since there are several ore passes in an underground mine, a human operator is responsible for the teleoperation of multiple impact hammers at once; however, a given operator can only control one machine at a time. The loaders that deposit material on the steel grates (e.g., load-haul-dump machines) cannot do so if the grates are completely obstructed due to large rocks, implying that the rock-breaking task is a potential bottleneck in the production chain.

The above poses the automation of impact hammers in underground mining as a relevant problem to solve due to the potential enhancements in terms of production that a fleet of efficient and autonomous impact hammers could provide. 

\subsubsection{Challenges and constraints}
\label{subsubsec:challenges_and_constraints}
Automating a fleet of impact hammers comes with many challenges. Firstly, many different models of impact hammers may be present in a mine, which means that developing a solution that accounts for different dynamics and control interfaces is required. Secondly, most impact hammers in a mine do not come with proprioceptive sensors other than those required to provide diagnostics on its hydraulic unit (e.g., sensors to measure the temperature and pressure of the hydraulic fluid of its tank), which is insufficient to implement classical closed-loop control systems. Finally, these machines are subjected to deterioration due to the task they perform, making replacement of their end-effectors, actuators, and other mechanisms common during maintenance sessions.

The aforementioned challenges impose requirements on any control system that is to be deployed on a fleet of impact hammers. The variability across impact hammers' models and the lack of sensor information on them motivate the selection of a minimal set of variables to characterize their state. This avoids relying on information that is not available or may not be obtained for all hammers in the fleet,\footnote{Examples on this matter include pressure measurements for a proper characterization of the hydraulic actuators' dynamics.} whilst standardizing and reducing implementation costs.

In addition, the variability across impact hammers' control interfaces has already resulted in the adoption of discrete control for their teleoperation in some underground mines. Although this design decision may prevent operators from having fine-grained control over the impact hammers, it comes with the benefit of only requiring characterizing the dead zones of the actuators of each impact hammer for its implementation at scale. This information is enough for the full standardization of the program that maps joystick commands to electro-valves setpoints for a whole fleet.\footnote{Note that this is only true if the machines composing said fleet share the same kinematic chain structure, which is generally the case for impact hammers with four degrees of freedom.}

\subsection{The Bobcat E10 mini-excavator}
\label{sec:bobcat_e10_miniexcavator}

As a testbed, we consider a Bobcat E10 mini-excavator equipped with an hydraulic impact hammer as end-effector. The hydraulic arm of this mini-excavator has the same kinematic chain structure as most impact hammers used in mining operations; in fact, this machine has previously been used to test an RL-based policy to control the impact hammer's boom and end-effector whenever the machine is in a suitable configuration to attempt breaking a rock~\citep{samtani2023learning}.

\begin{figure}
    \centering
    \includegraphics[width=\linewidth]{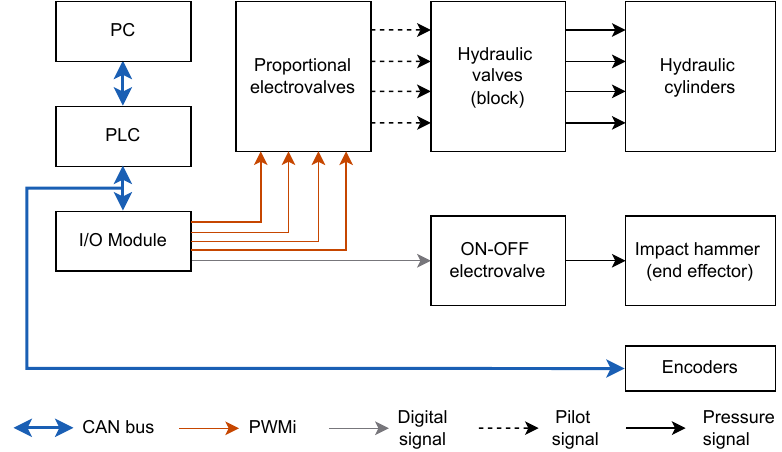}
    \caption{Simplified electro-hydraulic diagram for the modified Bobcat E10's mini-excavator.}
    \label{fig:bobcat_e10_simplified_hydraulics}
\end{figure}

The Bobcat E10 used in this work has been modified by installing electrovalves to control its actuators. To control its hydraulic arm, four pairs of electrovalves send pilot signals to the original machine's valves, so as to extend or retract the hydraulic cylinders of the arm. Similarly, to control the impact hammer used as end-effector, an on-off electrovalve is used. All of these electrovalves are actuated using a programmable I/O module connected to a programmable logic controller (PLC), which, in turn, is connected to an on-board PC and to a set of rotary encoders, using a CAN bus interface. The above is illustrated in Figure~\ref{fig:bobcat_e10_simplified_hydraulics}.

An important feature of the Bobcat E10 mini-excavator is that its arm's valves are integrated into a single hydraulic valve block, which makes it difficult to get accurate pressure measurements to monitor the state of the valves. This imposes restrictions on the overall control design for the arm, as there is missing information to fully characterize its dynamics. A simplified hydraulic circuit diagram of the intervened mini-excavator is described in Appendix~\ref{appendix:bobcat_e10_hydraulics}.

To get the Bobcat E10's hydraulic arm configuration, four absolute rotary encoders measure the angular position of each joint. Custom mounts were fabricated to accommodate each encoder so that their shafts are aligned to their corresponding joint axis. Moreover, these encoders are connected to the same PLC used for actuation, using the CAN protocol.

Although it may seem unusual to use rotary encoders to get the angular positions of the arm's joints instead of using draw-wire encoders to measure the displacement of each of its hydraulic cylinders, this design decision is due to two main factors related to impact hammers in underground mining. First, directly measuring joint positions using rotary encoders does not require a detailed modeling of the machine's kinematic chain (whereas draw-wire encoders generally do); moreover, for a given impact hammer in the fleet, the said kinematic chain may be unknown (for instance, because of structural modifications that occurred in maintenance sessions) and it will vary across different rock breaker models. Second, the use of properly protected draw-wire encoders, for instance those installed inside hydraulic cylinders, cannot be easily implemented in underground mines at scale, as it would require the disassembly of each intervened impact hammer and possibly the replacement of some actuators due to incompatibilities.

\section{Proposed approach}
\label{sec:proposed_approach}

To provide a solution to the problem of controlling a hydraulic impact hammer under limited observability and the requirement of using discrete control for its actuation, we follow an approach that can be described as a two-stage process. The first stage addresses the problem of modeling the impact hammers' dynamics (i.e., performing system identification), and the second stage addresses the problem of synthesizing a reaching controller for the machine, by leveraging the model obtained in the first stage. 

For the first stage, the parameters $\bm{\theta}$ of a function $f_{\bm{\theta}}$ that approximate the impact hammer's dynamics are learned through supervised learning given teleoperation data. For a history of previous (discrete) actions $(\bm{a}_{t-k}, ...\bm{a}_t)$, and arm configurations $(\bm{q}_{t-k}, ..., \bm{q}_t)$, the model $f_{\bm{\theta}}$ has to predict the next joint configuration, $\bm{q}_{t+1}$. 

For the second stage, the model $f_{\bm{\theta}}$ is used to generate experiences given some initial (approximate) state and a stream of actions, where the predicted configurations are fed back to the model itself to obtain rollouts over some temporal horizon. The generated data are used to obtain a reaching policy $\pi_{\bm{\phi}}$ via RL or MPC. Note that this policy is conditioned on observations $\bm{o}_t$ constructed using only the rollout data obtained using $f_{\bm{\theta}}$ to govern the underlying system dynamics. These observations contain both the current and target end-effector poses, $\mathcal{X}^{\text{eef}}_t$ and $\mathcal{X}^{\text{target}}$, respectively, where the current end-effector pose can be computed using forward kinematics. The synthesized policy can then be directly deployed in the real-world.

\begin{figure}
    \centering
    \includegraphics[width=\linewidth]{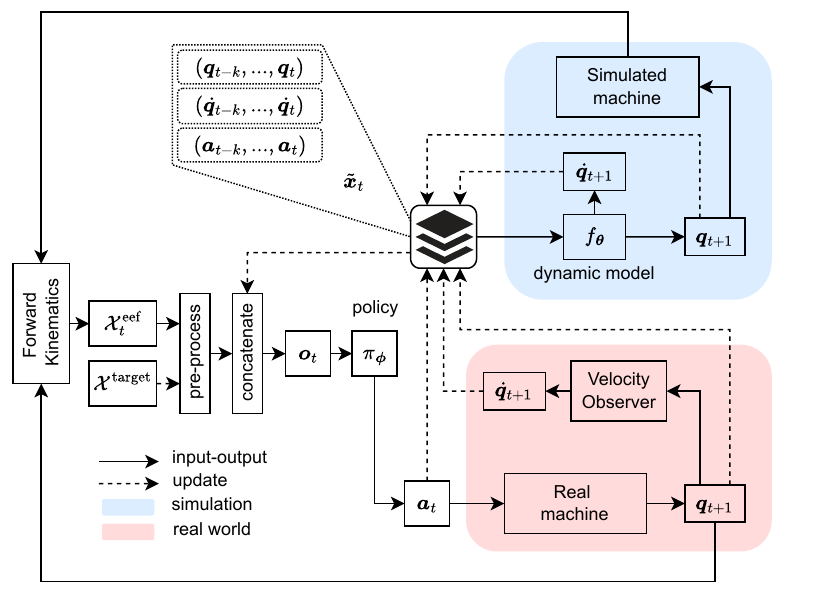}
    \caption{Overview of the methodology used to obtain reaching controllers for hydraulic impact hammers, given a learned dynamic model $f_{\bm{\theta}}$. Note that estimating $\dot{\bm{q}}_t$ is only necessary when $f_{\bm{\theta}}$ is trained to predict velocity residuals (see Section~\ref{sec:dynamic_model}).}
    \label{fig:system_overview}
\end{figure}

The whole described procedure is illustrated in Fig.~\ref{fig:system_overview}, where optionally, besides configurations and actions, joint velocity estimates, $\dot{\bm{q}}_{t}$, can be used to condition both the dynamic model $f_{\bm{\theta}}$ and the policy $\pi_{\bm{\phi}}$. In what follows, each stage of the proposed approach is described in detail.

\subsection{Data-driven system identification}
\label{subsec:data_driven_sysid}

We consider the problem of modeling the dynamics of a four-DoF hydraulic arm with an impact hammer attached as end-effector. To get a model of the machine kinematics, we must note that detailed 3D models of rock breakers, such as those used in mining, are rarely publicly available, and if they are, the real machines may not reflect the geometry documented by the manufacturer due to modifications performed during maintenance sessions. Therefore, the parameters of their kinematic chains, in general, must be obtained by taking measurements of the machines themselves (e.g, as in \cite{lampinen2021autonomous}), and then using CAD software to obtain coarse models of their links. The outcome of this process can then be utilized to specify an URDF file that describes the impact hammer's kinematics in a format that is ROS compatible~\citep{quigley2009ros}. However, in this work the arm's hydraulic cylinders are omitted given the limitations of the standard URDF when attempting to represent kinematic loops, and also because we use rotational encoders to directly measure the joints' angular positions. 

\begin{figure}
    \centering
    \includegraphics[width=\linewidth]{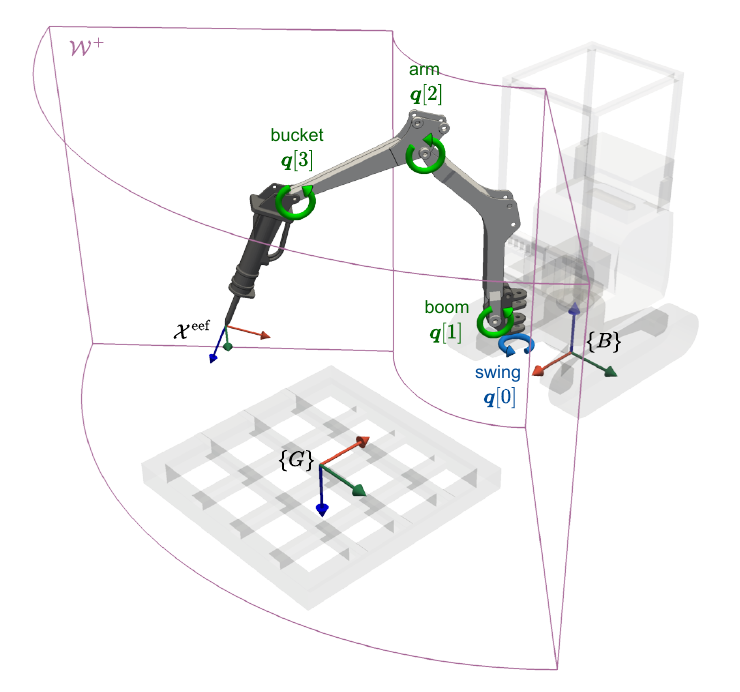}
    \caption{Kinematic layout of the Bobcat E10 mini-excavator, alongside a scaled steel grill. The fixed frames $\{B\}$ and $\{G\}$ are attached to the mini-excavator base link and to the center of the grill, respectively. The cylindrical sector defines the boundaries for the position of end-effector poses that may be elements of the restricted workspace $\mathcal{W}^+$.}
    \label{fig:bobcat_e10_kinematics}
\end{figure}

By applying the above on the Bobcat E10 mini-excavator, we get an URDF file that describes its kinematics. A rendering of the model obtained, alongside joint names, relevant frames, and a scaled steel grill similar to those used in real mining operations, is shown in Fig.~\ref{fig:bobcat_e10_kinematics}.

To model the response of the impact hammer's arm actuators given control commands, we follow a data-driven approach. However, many variables that would be required for an accurate dynamic model are often unknown, and we abstain from relying on measurements that would not be available straightforwardly for a fleet of impact hammers used in an underground mine. For the Bobcat E10  mini-excavator, examples for variables in the first category include pressure differentials for the valves (see Section~\ref{sec:bobcat_e10_miniexcavator}), whilst measurements that, although available, are purposefully omitted, include the RPMs of its motor and the temperature of its hydraulic fluid. 

Given the above, to learn a forward dynamics model for an impact hammer, and in particular, for the Bobcat E10's arm, we only consider the measurements provided by rotary encoders, and the control commands sent to electrovalves. Since not all rotary encoders provide speed estimations, we rely only on angular position measurements. These measurements are used to obtain the hydraulic arm configuration at time step $t$, $\bm{q}_t \in \mathbb{R}^4$. The electro-valve setpoints, on the other hand, are fixed due to the control restrictions for impact hammer fleets discussed in Section~\ref{subsubsec:challenges_and_constraints}. Thus, each hydraulic actuator contracts or expands given a fixed electrical signal, or may simply not act for a zero set point. Since these setpoints are fixed once tuned, a given (normalized) control command for time step $t$, $\bm{a}_t$, can be defined as an element of the set $\{-1, 0, 1\}^4$.

\subsubsection{Learning a model for the hydraulic arm actuators}
\label{sec:dynamic_model}

We aim to learn a model such that, given available measurements to characterize the state of the impact hammer for a discrete time step $t$, predicts what its configuration will be for the next time step $t+1$. We try to accomplish this by learning the parameters $\bm{\theta}$ of a model $f_{\bm{\theta}}$ that satisfies Eq.~\eqref{eq:general_model_form}, where $\bm{\tilde{x}}_t$ is a function of previous configurations and control commands.
\begin{equation}
    \bm{q}_{t+1} = \bm{q}_t+f_{\bm{\theta}}(\bm{\tilde{x}}_t)
    \label{eq:general_model_form}
\end{equation}

To properly characterize the state of the arm and the delays of the hydraulic actuators, $\bm{\tilde{x}}_t$ is defined according to Eq.~\eqref{eq:o_tilde}, where $\bm{q}_{t-k:t}=(\bm{q}_{t-k}, \bm{q}_{t-k+1}, ..., \bm{q}_t)\in \mathbb{R}^{4\times(k+1)}$,  $\bm{a}_{t-k:t}=(\bm{a}_{t-k}, \bm{a}_{t-k+1}, ..., \bm{a}_{t}) \in \mathbb{R}^{4\times(k+1)}$, and $g$ is the composition of functions that will be defined later.
\begin{equation}
    \bm{\tilde{x}}_t=g(\bm{q}_{t-k:t}, \bm{a}_{t-k:t})
    \label{eq:o_tilde}
\end{equation}

We consider two different approaches for a practical instantiation of parameterized models that satisfy Eq.~\eqref{eq:general_model_form}. For the first approach, the goal is to learn how to predict the residuals of the angular position ($\Delta \bm{q}_t$) to estimate $\bm{q}_{t+1}$. For the second approach, the goal is to learn how to predict angular velocity residuals ($\Delta \dot{\bm{q}}_t$) to obtain future velocity estimates and then compute $\bm{q}_{t+1}$ by integrating the predicted velocities. In both cases, multilayer perceptrons (MLPs) or Kolmogorov-Arnold networks (KANs)~\citep{liu2025kan} are utilized as function approximators. In what follows, each approach is described in further detail.

\begin{itemize}

    \item Learning $\Delta \bm{q}_t$: This approach follows Eq.~\eqref{eq:general_model_form} by representing $f_{\bm{\theta}}$ as an MLP or a KAN, and defining $g$ as the composition of a min-max normalization for the arm configurations, a flattening operation over both configurations and actions, and the concatenation of the min-max normalized current configuration to the result. The min-max normalization for the angular positions of the arm is defined by Eq.~\eqref{eq:min_max_normalization}, where the $j$-th components of $\bm{q}_{\text{min}}$ and $\bm{q}_{\text{max}}$  are set according to the joint limits of the arm, and $\bm{1}$ is a column vector in $\mathbb{R}^4$ with all its components equal to $1$.
    \begin{equation}
        \texttt{minmax}(\bm{q}_t, \bm{q}_{\text{min}}, \bm{q}_\text{max}) =2\left(\frac{\bm{q}_t-\bm{q}_{\text{min}}}{\bm{q}_\text{max}-\bm{q}_\text{min}}\right) - \bm{1} = \bar{\bm{q}}_t
        \label{eq:min_max_normalization}
    \end{equation}
    
    The above results in $\bm{\tilde{x}}_t$ being defined by Eq.~\eqref{eq:x_t_pos}\footnote{Note that, although $\bm{\bar{q}}^T_{t}$ is already in $\bm{\bar{q}}^T_{t-k:t}$, we include it as the last four components of $\bm{\tilde{x}}_t$ in Eq.~\eqref{eq:x_t_pos} to get vectors of the same dimensions and overall structure than those defined by Eq.~\eqref{eq:x_t_vel}. In practice, this redundancy is introduced to simplify implementation.}, where $\bm{\bar{q}}^T_{t-k:t}=(\bm{\bar{q}}_{t-k}^T, \bm{\bar{q}}_{t-k+1}^T, ..., \bm{\bar{q}}_t^T)\in \mathbb{R}^{4(k+1)}$ and $\bm{a}^T_{t-k:t}=(\bm{a}_{t-k}^T, \bm{a}_{t-k+1}^T, ..., \bm{a}_t^T)\in \mathbb{R}^{4(k+1)}$ are row vectors.
    \begin{align}
        \bm{\tilde{x}}_t = (\bm{\bar{q}}^T_{t-k:t}, \bm{a}^T_{t-k:t}, \bm{\bar{q}}^T_{t})\in\mathbb{R}^{8(k+1)+4}
        \label{eq:x_t_pos}
    \end{align}

    \item Learning $\Delta \dot{\bm{q}}_t$: This approach follows Eq.~\eqref{eq:general_model_form} by defining $f_{\bm{\theta}}(\bm{\tilde{x}}_t):=\Delta t\cdot(\dot{\bm{q}}_{t-1}+f'_{\bm{\theta}}(\bm{\tilde{x}}_t))$, such that $f'_{\bm{\theta}}$ (represented by an MLP or a KAN), predicts velocity residuals to get velocity estimations that are then integrated to get future angular positions, as in Eqs. \eqref{eq:vel_prediction} and \eqref{eq:pos_prediction}.
    \begin{align}
         \dot{\bm{q}}_{t} &= \dot{\bm{q}}_{t-1} + f'_{\bm{\theta}}(\bm{\tilde{x}}_t)  \label{eq:vel_prediction}\\
         \bm{q}_{t+1} &= \bm{q}_t + \Delta t \cdot \dot{\bm{q}}_{t}  \label{eq:pos_prediction}
    \end{align}

    Here, $f'_{\bm{\theta}}(\bm{\tilde{x}}_t)$ is conditioned on velocities which are estimated given measured angular positions. To get these estimates, a loop tracker for the measured joint positions is implemented, such that a filtered angular velocity estimation can be obtained as an intermediate variable. For parameters $k_i$ and $k_p$, the implemented loop tracker for a given joint is illustrated in Fig.~\ref{fig:loop_tracker}. Note that to filter high frequencies, we do not use $\hat{\dot{q}}$ as the velocity estimate; instead, we use the integral term of the proportional-integral (PI) controller that tracks the joint position.
  
    Using the loop tracker, we can get an estimate for $\dot{\bm{q}}_t$ given a history of joint positions up to time step $t$. Similarly to the case where we learn angular position residuals, here we define $g$ as the composition of a velocity estimator (implemented as previously described), a min-max normalization of angular velocity estimations (as $\texttt{minmax}(\dot{\bm{q}}_t, \dot{\bm{q}}_{\text{min}}, \dot{\bm{q}}_\text{max})$, see. Eq.~\eqref{eq:min_max_normalization}), a flattening operation over velocity estimations and actions, and the concatenation of the min-max normalized current configuration. This results in $\bm{\tilde{x}}_t$ being defined by Eq.~\eqref{eq:x_t_vel}.
    \begin{align}
        \bm{\tilde{x}}_t = (\bar{\dot{\bm{q}}}^T_{t-k:t}, \bm{a}^T_{t-k:t}, \bar{\bm{q}}^T_{t})\in\mathbb{R}^{8(k+1)+4}
        \label{eq:x_t_vel}
    \end{align}

    \begin{figure}
        \centering
        \includegraphics[width=\linewidth]{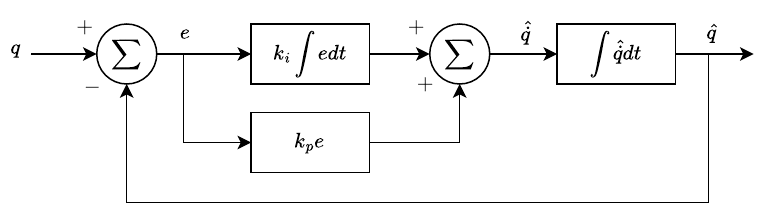}
        \caption{Block diagram of the loop tracker implemented to get $\dot{\bm{q}}_t$ estimates.}
        \label{fig:loop_tracker}
    \end{figure}
 
\end{itemize}

The models that predict residuals are trained using backpropagation through time \citep{werbos1988generalization}. To do so, we rely on a dataset $\mathcal{D}$ of temporally ordered tuples $(\bm{q}_t, \dot{\bm{q}}_t, \bm{a}_t)$ that allows the construction of short trajectories of the form ${\tau_i=(\bm{q}_{t_i}, \dot{\bm{q}}_{t_i}, \bm{a}_{t_i}, ..., \bm{q}_{t_{i}+H}, \dot{\bm{q}}_{t_{i}+H}, \bm{a}_{t_{i}+H})}$ for a given initial time step $t_{i}$, along with a corresponding state approximation $\tilde{\bm{x}}_{t_{i}}$ associated with the first configuration action tuple of the trajectory. These trajectories are used to calculate a loss function that measures the approximation error of the models over a short time horizon~$H$:
\begin{equation}
    \mathcal{L}(\bm{\theta})= \frac{1}{H}\sum_{j=1}^H\|\hat{\bm{q}}_{t_{i}+j} - \bm{q}_{t_{i}+j}\|^2_2,
    \label{eq:dmodel_loss_function}
\end{equation}

where the configurations $\hat{\bm{q}}_{t_{i}+j}$ are predicted in an ``open-loop'' manner by the model, that is, starting from  $\tilde{\bm{x}}_{t_{i}}$:
\begin{align*}
     \hat{\bm{q}}_{t_{i}+1}&=\bm{q}_{t_i}+f_{\bm{\theta}}(\tilde{\bm{x}}_{t_i}),\\
     \hat{\bm{q}}_{t_{i}+2}&=\hat{\bm{q}}_{t_i+1}+f_{\bm{\theta}}(\hat{\tilde{\bm{x}}}_{t_i+1}),\\
     & \vdotswithin{=}\\
     \hat{\bm{q}}_{t_{i}+H}&=\hat{\bm{q}}_{t_i+H-1}+f_{\bm{\theta}}(\hat{\tilde{\bm{x}}}_{t_i+H-1}).\\
\end{align*}
Note that $\tilde{\bm{x}}_{t_i+j}$, $j\in\{1,...,H-1\}$ is constructed by progressively updating the components of $\tilde{\bm{x}}_{t_i}$ with the ground-truth actions in $\tau_i$ and the predictions of the model (which may be angular positions or velocities, depending on the chosen model).

\subsection{Controller synthesis for reaching}

\subsubsection{Problem formulation}
\label{subsubsec:problem_formulation}

In this work, we address the reaching task for hydraulic rock breakers, that is, we aim at synthesizing a controller capable of reaching end-effector target poses. In practice, performing this task would allow these machines to position their hydraulic hammer on top of rocks that need to be fractured.  

The interaction between the agent and the environment is modeled as a Partially Observable Markov Decision Process (POMDP), which is defined by a set of states $\mathcal{S}$, a set of actions $\mathcal{A}$, a scalar reward function $\mathcal{R}(\bm{s},\bm{a})$, a stochastic transition function $T(\bm{s},\bm{a},\bm{s}')=p(\bm{s}'|\bm{s},\bm{a})$, an observation function $\mathcal{O}(\bm{s}', \bm{a}, \bm{o}) = p(\bm{o}|\bm{s}',\bm{a})$, a set of observations $\Omega$, and a discount factor $\gamma \in [0, 1)$. At each discrete time step $t$ the agent observes $\bm{o}_t$, executes an action $\bm{a}_t$ according to its policy $\pi(\bm{a}_t|\bm{o}_t)$, receives a scalar reward $r_t$, and transitions to a new state $\bm{s}_{t+1}$. 

We will assume that the rock breakers operate in a restricted region of their workspace and that this region is always free of obstacles (see Section~\ref{subsec:episodic_conditions}). This allows solving the reaching task without relying on exteroceptive information.

\subsubsection{Modeling}
\label{subsubsec:modeling}
\begin{itemize}

    \item \textbf{Dynamics:} The impact hammer dynamics is forward simulated using a parameterized function $f_{\bm\theta}$, which is trained as described in Section~\ref{sec:dynamic_model}. Depending on whether the learned model predicts the angular position residuals or the angular velocity residuals, this is accomplished by iteratively evaluating Eq.~\eqref{eq:general_model_form}, or Eqs.~\eqref{eq:vel_prediction} and~\eqref{eq:pos_prediction}.

    \item \textbf{Observations:} To provide the agent with enough information to make decisions (given the underlying learned dynamics), the policy is conditioned on $\tilde{\bm{x}}_t$, on a representation of the current impact hammer's end-effector pose, $\mathcal{X}^\text{eef}_t \in \text{SE(3)}$, and on a representation of the target pose, $\mathcal{X}^\text{target} \in \text{SE(3)}$. Note that $\tilde{\bm{x}}_t$ is given by Eq.~\eqref{eq:x_t_pos} or by Eq.~\eqref{eq:x_t_vel}, depending on the chosen dynamics model.
    
    We use two complementary representations for the end effector poses (instantaneous and target):
    \begin{itemize}
        \item As tuples $(\bm{p}, \bm{r})$, where $\bm{p}\in \mathbb{R}^3$ denotes position and $\bm{r}\in \mathcal{S}^3$ denotes orientation.
        \item As configurations $\bm{q}\in \mathbb{R}^4$ that would be mapped to the poses via forward kinematics.
    \end{itemize}
    
    Thus, the observations $\bm{o}_t$ are given by the concatenation of $\tilde{\bm{x}}_t$ with representations for the instantaneous and target end-effector poses. 
    
    \item \textbf{Actions:} 
    Given the operational constraints described in Section~\ref{subsubsec:challenges_and_constraints}, the control of the impact hammer is done using discrete control commands. Thus, using the same definition as in Section~\ref{subsec:data_driven_sysid}, a normalized action at time $t$, $\bm{a}_t$, is an element of the set $\{-1,0, 1\}^4$.
    
    \item \textbf{Reward function:} The reward function is designed to encourage the completion of the reaching task, while also regularizing the policy's actions to prevent the exploitation of the underlying learned dynamics inaccuracies.

     We define a penalty function to measure the deviation of the current end-effector pose, $\mathcal{X}^{\text{eef}}_t$, from the target pose, $\mathcal{X}^{\text{target}}$, by weighting their positional and rotational differences. Representing the poses as tuples $(\bm{r}, \bm{p})$, the position error is quantified as the Euclidean distance between their positions:
    \begin{equation}  
    d(\bm{p}^{\text{target}},\bm{p}_t^{\text{eef}} ) = \| \bm{p}^{\text{target}} - \bm{p}_t^{\text{eef}} \|_2,
    \end{equation}
    
    whereas the rotational error is computed as the normalized geodesic distance between their unit quaternions:
    \begin{equation}    
    d_g(\bm{r}^{\text{target}}, \bm{r}_t^{\text{eef}}) = \frac{1}{\pi} \cos^{-1}\left( 2 (\bm{r}^{\text{target}} \cdot \bm{r}_t^{\text{eef}})^2 - 1 \right).
    \end{equation}

    Thus, the described penalization function is given by:
    \begin{equation}
        r^{\mathcal{X}}_t = -\lambda_{\bm{p}} d(\bm{p}^{\text{target}},\bm{p}_t^{\text{eef}} ) - \lambda_{\bm{r}} d_g(\bm{r}^{\text{target}}, \bm{r}_t^{\text{eef}}),
    \end{equation}
    where $\lambda_{\bm{p}}, \lambda_{\bm{r}}>0$ are fixed scalars.
    
     In addition, we include a penalty for deviations in joint-space from the desired target configuration, $\bm{q}^{\text{target}}$ (which can be mapped to $\mathcal{X}^{\text{target}}$ via forward kinematics):
    \begin{equation}
        r_t^{\bm{q}} = - \lambda_{\bm{q}}\| \bm{q}_t - \bm{q}^{\text{target}} \|_2.
    \end{equation}
    
    To encourage fine-grained control, we consider binary rewards that the agent receives only if it manages to meet certain accuracy criteria. These binary rewards are computed in terms of positional and rotational errors (Eqs.~\eqref{eq:b_pose}\footnote{Here $\llbracket \cdot \rrbracket$ is the Iverson bracket, meaning $\llbracket P \rrbracket=1$ if $P$ is True, and $\llbracket P \rrbracket=0$ otherwise.} and~\eqref{eq:b_angle}\footnote{We only consider the yaw and pitch Euler angles, since given the impact hammer kinematics, the roll-angle is fixed. Although a similar argument could be used to also omit the yaw-angle, we use it because its value is fully determined by the ``swing'' joint, i.e, by $\bm{q}_t[0]$ (see Fig.~\ref{fig:bobcat_e10_kinematics}).}), and deviations in joint space (Eq.~\eqref{eq:b_q}), where $\epsilon_{\bm{p}}, \epsilon_{\bm{r}}, \epsilon_{\alpha}, \epsilon_q>0$ are fixed scalar thresholds.
    \begin{align}
        b_\mathcal{X} &= \bigl\llbracket d(\bm{p}^{\text{target}},\bm{p}_t^{\text{eef}})<\epsilon_{\bm{p}})\bigr \rrbracket \cdot \bigl\llbracket(d_g(\bm{r}^{\text{target}}, \bm{r}_t^{\text{eef}})<\epsilon_{\bm{r}})\bigr \rrbracket \label{eq:b_pose}\\
        b_{\alpha} &=\llbracket(\alpha_{\text{yaw}} - \alpha_{\text{yaw}}^{\text{target}})^2<\epsilon_\alpha \rrbracket \cdot \llbracket(\alpha_{\text{pitch}} - \alpha_{\text{pitch}}^{\text{target}})^2<\epsilon_\alpha \rrbracket \label{eq:b_angle}\\
       b_{\bm{q}} &=\Pi_{j=0}^3\llbracket(\bm{q}_t[j] - \bm{q}^{\text{target}}[j])^2<\epsilon_q \rrbracket \label{eq:b_q}
    \end{align}
    The reward the agent receives depending on the fulfillment of the above conditions is given by Eq.~\eqref{eq:accuracy_bonus}, where $w_\mathcal{X},  w_{\alpha}, w_{\bm{q}}>0$ are fixed scalars.
    \begin{equation}
        r^{\epsilon}_t = w_\mathcal{X}\left(b_\mathcal{X}|_{(\epsilon_{\bm{p}},\epsilon_{\bm{r}})}+b_{\mathcal{X}}|_{(\epsilon'_{\bm{p}},\epsilon'_{\bm{r}})}\right) + w_{\alpha}b_{\alpha} + w_{\bm{q}}b_{\bm{q}}
        \label{eq:accuracy_bonus}
    \end{equation}

    To encourage temporal consistency in the selection of actions, we penalize changes in action components between successive time steps:
    \begin{equation}
    r_t^{\bm{a}} = -\lambda_{\bm{a}} \cdot \frac{1}{4} \sum_{i=0}^{3} |\bm{a}_t[i] - \bm{a}_{t-1}[i]|,
    \end{equation}
    where $\lambda_{\bm{a}} > 0$ is the action regularization weight.

    Finally, we also penalize the agent if the impact hammer's end effector goes outside a restricted region of its workspace, $\mathcal{W}^{+}\subset\mathcal{W}\subset \text{SE}(3)$ (see Fig.~\ref{fig:bobcat_e10_kinematics}), where $\mathcal{W}$ denotes the full reachable workspace, and $\lambda_{\mathcal{W}^{+}}>0$ is a fixed scalar:
    \begin{equation}
         r_t^{\mathcal{W}^{+}} = -\lambda_{\mathcal{W}^{+}}\llbracket\neg(\mathcal{ X}_t^{\text{eef}}\in\mathcal{W}^ {+})\rrbracket.
    \end{equation}

    Thus, the full reward is a weighted combination of all of the above terms:
    \begin{equation}
    r_t = r_t^{\mathcal{X}} + r_t^{\bm{q}} + r_t^{\epsilon} + r_t^{\bm{a}} + r_t^{\mathcal{W}}.
    \label{eq:reward_function}
    \end{equation}

\end{itemize}

\subsubsection{Episodic conditions}
\label{subsec:episodic_conditions}

The reaching task, as formulated in this work, is episodic. In each episode, a random initial configuration for the impact hammer is selected. This initial configuration is such that the position of the end-effector of the hammer remains inside a restricted region of its workspace, which is denoted by $\mathcal{W^{+}}$ (see Fig.~\ref{fig:bobcat_e10_kinematics}), and its orientation has a bounded pitch angle, $|\alpha_\text{pitch}^{\text{target}}|\leq\alpha_{\text{pitch}}^{\text{max}}$. Afterwards, a target end effector pose, $\mathcal{X}_\text{target}\in \text{SE}(3)$, also within the restricted workspace and satisfying the pitch angle condition, is randomly selected.

For a given episode, if the end effector of the impact hammer reaches $\mathcal{X}_\text{target}$, then the episode is deemed successful. On the contrary, if the hammer's end effector does not reach $\mathcal{X}_\text{target}$ after $T_\text{reset}=500$ discrete time steps, then the episode is considered unsuccessful, and episodic conditions are reset. Note that these conditions are checked only to measure the performance of the reaching controller but do not imply getting to a terminal state.

\begin{figure}
    \centering
    \subfloat[][]{\includegraphics[width=\linewidth]{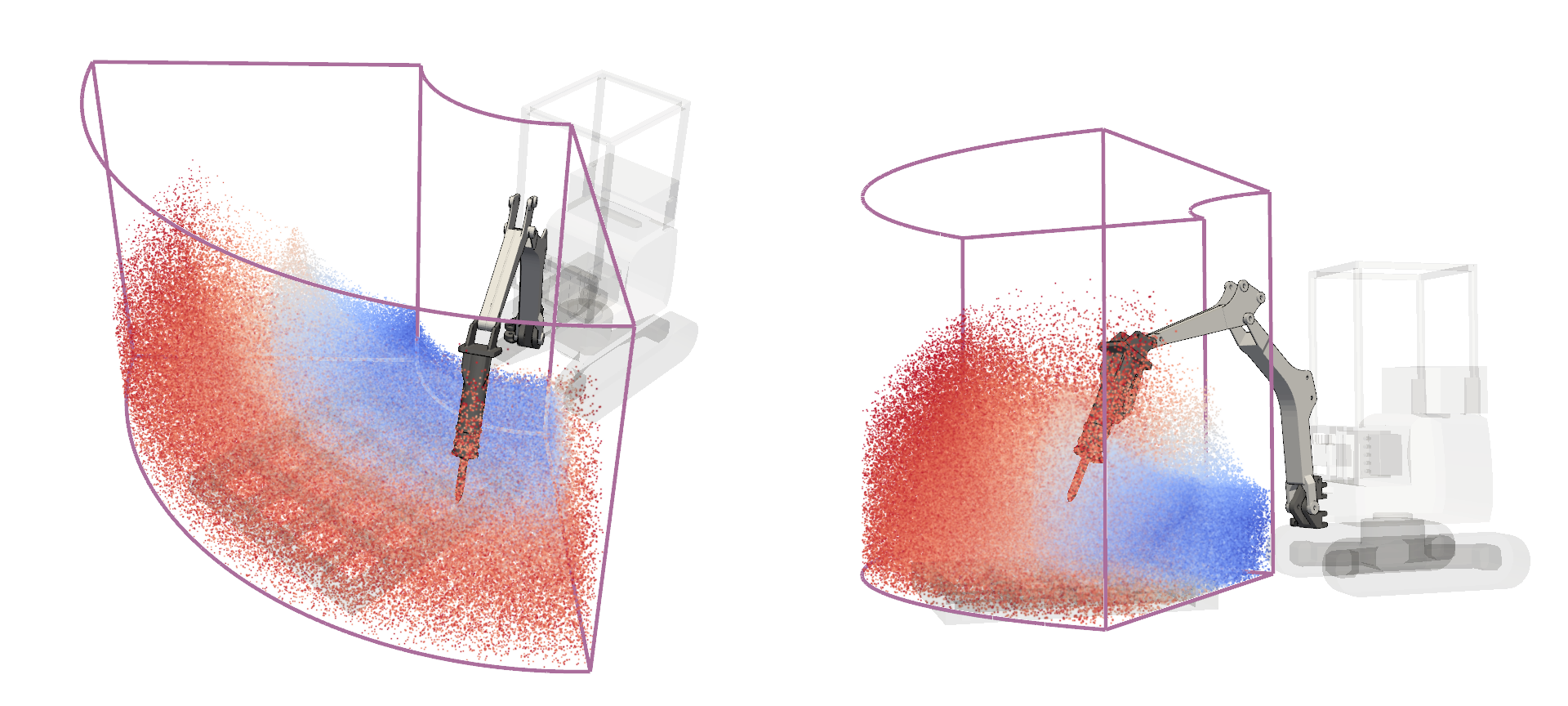}\label{fig:ws_visualization}} \\
    \subfloat[][]{\includegraphics[width=\linewidth]{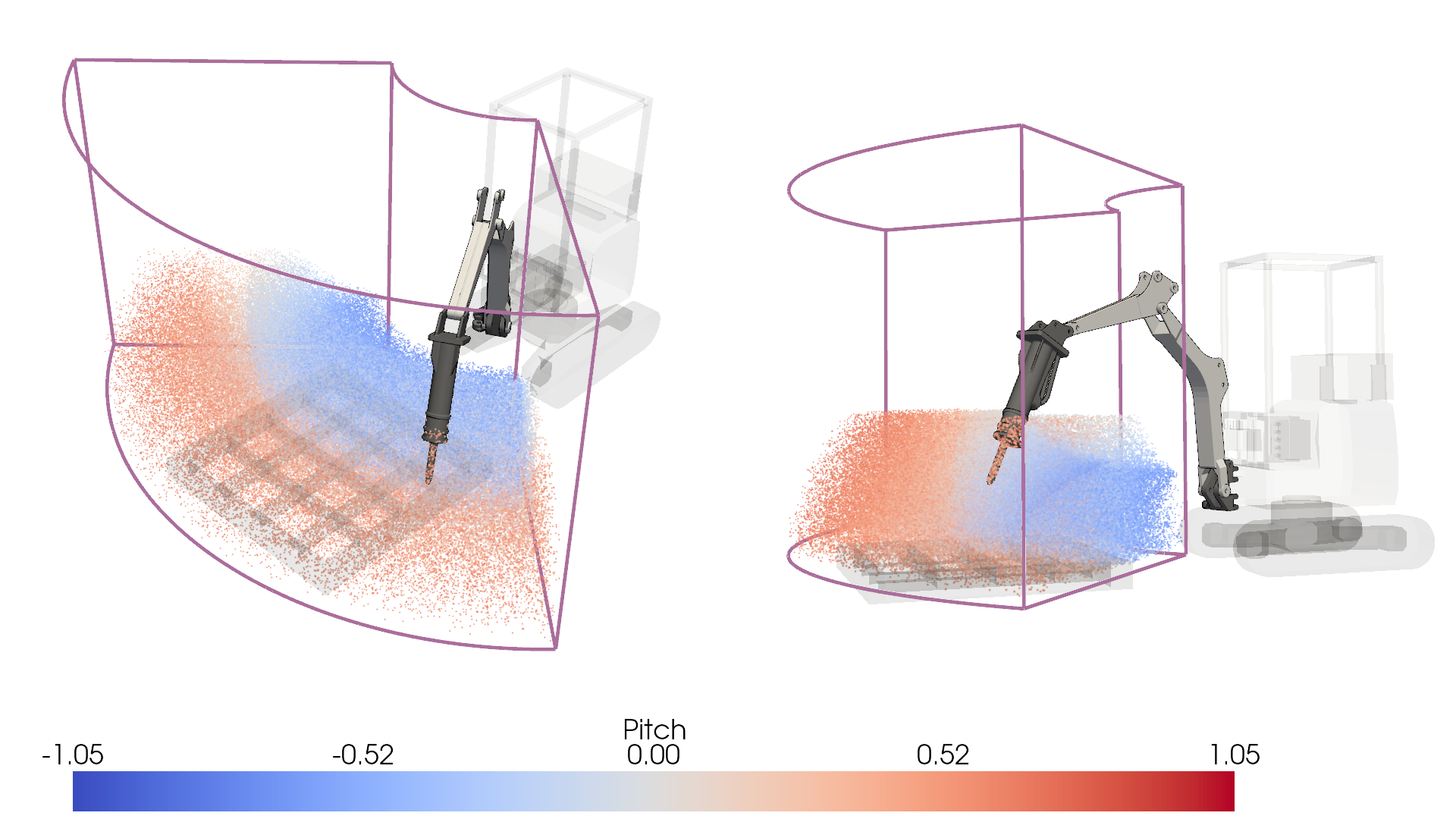}\label{fig:ts_visualization}} 
    \caption{Sampled end-effector poses used to set episodic conditions for the reaching task during (a) policy learning, and (b) policy evaluation.}
    \label{fig:ws_ts_visualization}
\end{figure}

Fig.~\ref{fig:ws_visualization} shows sampled end-effector poses that fulfill the position and pitch conditions described above for the Bobcat~E10 mini-excavator, by setting $\alpha_{\text{pitch}}^{\text{max}}=1.05$ [rad]. Fig.~\ref{fig:ts_visualization}, on the other hand, shows the sampled end effector poses that, besides the above conditions, fulfill $|\alpha_\text{pitch}^{\text{target}}|\leq\alpha_{\text{pitch}}^{\text{ts}}$, with $\alpha_{\text{pitch}}^{\text{max}} >\alpha_{\text{pitch}}^{\text{ts}}=0.69$ [rad], and the $z$-axis component of their position being in $[z^{\text{ts}}_\text{min}, z^{\text{ts}}_\text{max}]$, with $z^{\text{ts}}_\text{min}=0.155$ [m] (which matches the grill height) and $z^{\text{ts}}_\text{max}=2.355$ [m], respectively.

During policy learning, initial and target poses (and their respective configurations) are sampled considering the conditions applied to generate Fig.~\ref{fig:ws_visualization}. For evaluation purposes the same applies for initial poses, however, the target poses are obtained by enforcing the conditions used to generate Fig.~\ref{fig:ts_visualization}. Note that the latter poses, besides being a subset of those used during learning, fulfill restrictions that are \textit{ad-hoc} to the task space associated to rock-breaking, in which it is expected for the end-effector to reach poses that are quasi-orthogonal to the surfaces of the rocks that need to be fractured.

\subsubsection{RL-based controller}
\label{subsubsec:rl_controllers}

Given the formulation described in Section~\ref{subsubsec:problem_formulation}, the goal is to learn a policy that maximizes the expected discounted return that the agent receives, i.e., to maximize $J_{\text{RL}}(\pi)=\mathbb{E}_{\bm{a}_t\sim \pi(\bm{a}_t|\bm{o}_t)}\left[\sum_{t=1}^T \gamma^{t-1}r_t\right]$. 

To accomplish the above, we train a policy using the Proximal Policy Optimization (PPO) algorithm~\citep{schulman2017proximal}. To adapt PPO to discrete actions, we follow an approach close to that proposed by~\cite{dulac2015deep} for their ``Wolpertinger architecture'': Given a continuous proto-action $\bm{a}^p_t\in[-1,1]^4$ (output by the PPO policy), we construct a valid action $\bm{a}_t$ by discretizing each component of the proto-action, $\bm{a}_t^p[j], j\in\{0,1,2,3\}$, according to Eq.~\eqref{eq:proto_discretization}.

\begin{equation}
  \bm{a}_t[j] =
    \begin{cases}
        -1 & \text{if } \bm{a}^p_t[j] < -0.5, \\
        0 & \text{if } \left|\bm{a}^p_t[j]\right| \leq 0.5, \\
        1 & \text{if } \bm{a}^p_t[j] > 0.5
   \end{cases}
   \label{eq:proto_discretization}
\end{equation}

While the discretized action $\bm{a}_t$ is used for control, the proto-action is used to update the policy given the learning objectives of PPO.
    
\subsubsection{MPC-based controller}
\label{subsubsec:mpc_controllers}

As an alternative to RL, we also synthesize a controller following an MPC approach. MPC formulates control as a receding horizon optimization problem that leverages the learned dynamic model to plan action sequences over a finite time horizon. At each time step $t$, the MPC controller solves the following optimization problem:
\begin{align}
    \bm{a}^*_{t:t+H_{\text{MPC}}-1} &= \argmax_{\bm{a}_{t:t+H_{\text{MPC}}-1}} \sum_{k=0}^{H_{\text{MPC}}-1} \gamma^k r_{t+k} \label{eq:mpc_objective}\\
    \text{subject to:} \: & \nonumber \\
    \bm{q}_{t+k+1} &= \bm{q}_{t+k}+f_{\bm{\theta}}(\bm{\tilde{x}}_{t+k}), \quad k \in \{0, ..., H_{\text{MPC}}-1\} \label{eq:mpc_dynamics}\\
    \bm{a}_{t+k} &\in \{-1, 0, 1\}^4, \quad  k \in \{0, ..., H_{\text{MPC}}-1\} \label{eq:mpc_discrete}
\end{align}
where $\bm{a}^*_{t:t+H_{\text{MPC}}-1}$ denotes the optimal action sequence over the planning horizon $H_{\text{MPC}}$, $\gamma$ is the discount factor, and $r_{t+k}$ are the rewards computed using the function defined in Section~\ref{subsubsec:modeling}. Constraint~\eqref{eq:mpc_dynamics} enforces that the predicted configurations evolve according to the learned dynamic model $f_{\bm{\theta}}$, and constraint~\eqref{eq:mpc_discrete} restricts the actions to be discrete. Note that only the first action $\bm{a}^*_t$ of the optimal sequence is executed, and the optimization is repeated at the next time step with updated state information.

To solve the problem defined by Eqs.~\eqref{eq:mpc_objective}--\eqref{eq:mpc_discrete}, we employ a variant of the Cross-Entropy Method (CEM)~\citep{de2005tutorial}, a sampling-based optimization algorithm that iteratively refines a distribution over candidate action sequences. At each iteration, CEM samples a population of $N_{\text{pop}}$ continuous proto-action sequences $\{\bm{a}^{p,(i)}_{t:t+H_{\text{MPC}}-1}\}_{i=1}^{N_{\text{pop}}}$ from a parameterized Gaussian distribution $\mathcal{N}(\bm{\mu}, \bm{\Sigma})$ with diagonal covariance matrix over the continuous action space $[-1, 1]^{4\times H_{\text{MPC}}}$, discretizes each proto-action to obtain valid discrete action sequences $\{\bm{a}^{(i)}_{t:t+H_{\text{MPC}}-1}\}_{i=1}^{N_{\text{pop}}}$ according to Eq.~\eqref{eq:proto_discretization}, evaluates their cumulative rewards using the learned model $f_{\bm{\theta}}$ via forward simulation, selects the top $N_{\text{elite}}$ sequences with highest returns, and updates the distribution parameters $(\bm{\mu}, \bm{\Sigma})$ to concentrate probability mass around these elite samples. This process repeats for a fixed number of iterations or until convergence, yielding an approximately optimal action sequence. 

The variant we use is referred to as iCEM (improved CEM), which incorporates the enhancements proposed by~\cite{pinneri2021sample} into the standard CEM formulation. These improvements include colored action noise to enforce temporal smoothness in the planned trajectories, importance mixing to leverage information from previous optimization rounds, and momentum-based updates to stabilize the distribution refinement process. These modifications have been shown to improve sample efficiency and solution quality, particularly in contact-rich manipulation tasks with complex dynamics \citep{Sancaktar2022Curious, Li2024DeformNet, Jiang2024DexSim2Real2}.

\section{Experimental results}
\label{sec:experimental_results}

\subsection{Computational pipelines}
\label{subsec:computational_pipeline}

The models used for the data-driven identification of the system (see Section~\ref{subsec:data_driven_sysid}) are implemented and trained using Flax~\citep{flax2020github}. The simulation environment is entirely built on top of JAX~\citep{jax2018github}, taking advantage of XLA (accelerated linear algebra), just-in-time (JIT) compilation, and parallelization. Moreover, the simulation pipeline integrates Brax~\citep{brax2021github} to compute forward kinematics. For visualization purposes, we leverage the data structures provided by Brax to render the simulation using the MuJoCo~\citep{todorov2012mujoco} visualizer. For RL-based policy learning, we employ the PPO implementation provided by Brax, whereas for MPC-based control, iCEM is implemented in JAX. 

Implementing simulation, modeling, and learning in JAX (and JAX-based libraries) allows for a highly efficient computational workflow. As a result, for instance, we can train RL-based policies using PPO for hundreds of millions of steps in just a few minutes on mid-range consumer desktops (e.g., equipped with an Intel i7-12700 CPU, an NVIDIA RTX 4060 GPU, and 32 GB of RAM).

To deploy the controllers on the real machine, a ROS-based pipeline~\citep{quigley2009ros} was developed. Moreover, a pre-deployment stage using the Gazebo simulator~\citep{koenig2004design} was incorporated to allow testing in a safe environment, thereby minimizing the risk of accidents that could harm either staff or the machine. The implementation details for the ROS-based and ROS-Gazebo computational pipelines are provided in the Appendix~\ref{appendix:ros_based_pipelines}.

\subsection{Data collection for system identification}
\label{subsec:data_collection}

We collect teleoperation data to construct a dataset $\mathcal{D}$ of temporally ordered tuples $(\bm{q}_t, \dot{\bm{q}}_t, \bm{a}_t)$. This dataset is then used to learn the parameters of the dynamic model $f_{\bm{\theta}}$ through the minimization of the loss function in Eq.~\eqref{eq:dmodel_loss_function} (see Sec.~\ref{sec:dynamic_model} for a detailed explanation). 

The teleoperation of the Bobcat E10 mini-excavator is performed by sending discrete commands $\bm{a}_t \in \{-1, 0, 1\}^4$ to its actuators. These control commands are constructed by discretizing the analog signals from the left and right sticks of a wireless Xbox 360 controller. The joystick interpreter is implemented in ROS, and the control commands are sent via CAN bus to the PLC installed in the mini-excavator. The PLC program performs a direct mapping of the discrete control commands to current set points for each electro-valve, so each stick axis of the Xbox 360 controller (four in total) can actuate one of the four hydraulic cylinders of the mini-excavator arm. It is important to note that this direct joint-level teleoperation resembles that of impact hammers in underground mining (see Section~\ref{subsubsec:challenges_and_constraints}).

The instantaneous angular positions of the hydraulic arm joints, $\bm{q}_t$, are obtained directly through rotational encoder measurements (as explained in Section~\ref{sec:bobcat_e10_miniexcavator}), and their angular velocity, $\dot{\bm{q}}_t$, is estimated using the observer described in Section~\ref{sec:dynamic_model}.

Using the ROS-based teleoperation module, the mini-excavator arm is controlled at 20 Hz, and the tuples $(\bm{q}_t, \dot{\bm{q}}_t, \bm{a}_t)$ are stored at the same rate. To construct a dataset with representative samples, the mini-excavator arm is operated with the objective of covering the subset of its configuration space that would result in its end-effector position to lie within the restricted workspace, $\mathcal{W}^{+}$, described in Section~\ref{subsec:episodic_conditions}. With this aim, while controlling the machine, the human operator can see a 3D representation of the restricted workspace limits, and has real-time visual feedback of the history of visited end-effector positions, thus, is instructed to ``paint'' the interior of the restricted workspace, with the end effector position acting as a 3D pencil.

\begin{figure*}
    \centering
    \includegraphics[width=\linewidth]{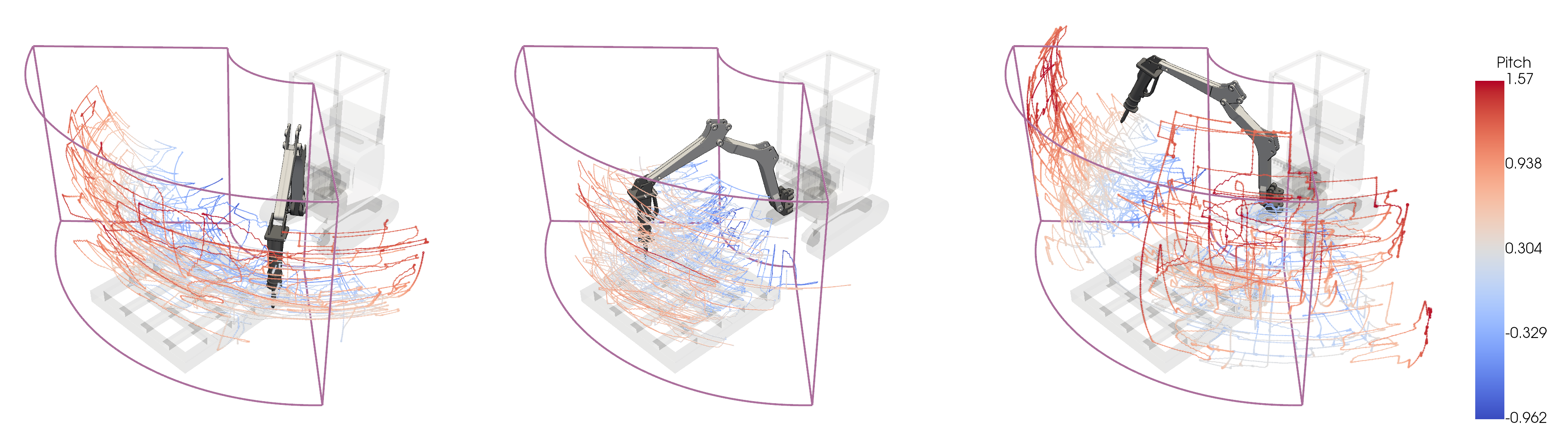}
    \caption{End effector trajectories resulting from the teleoperation of the Bobcat E10 mini-excavator, for three data collection sessions (left, center, and right). The trajectories' color represent the instantaneous end effector pitch angle.}
    \label{fig:db_trajectories}
\end{figure*}

Fig.~\ref{fig:db_trajectories} shows the visual feedback described, constructed from data acquired by teleoperating the mini-excavator arm for approximately 21, 23 and 32 minutes, respectively.

\begin{figure*}[t]
    \centering
    \includegraphics[width=\linewidth]{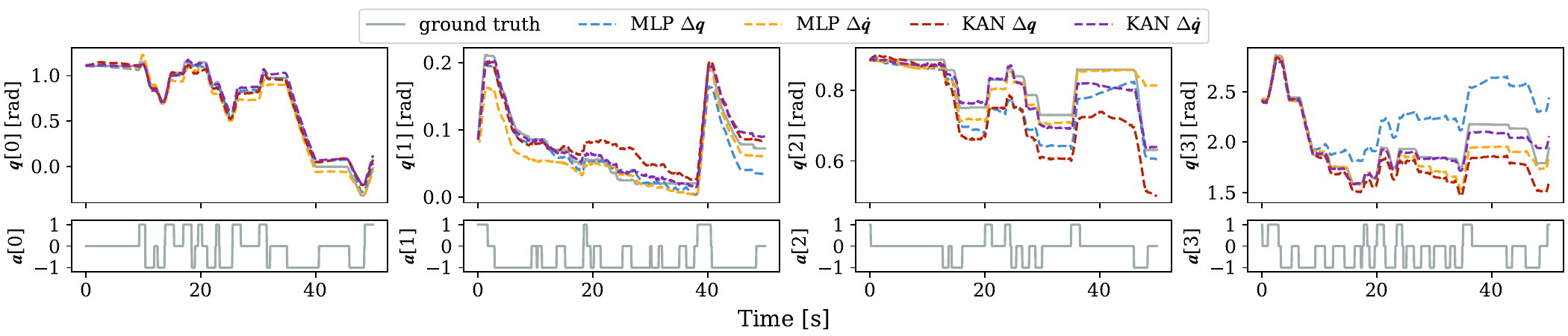}
    \caption{Open-loop predictions performed by the best learned dynamic models over an evaluation subset. Given an initial ground-truth construction of $\tilde{\bm{x}}_t$ and a sequence of $10^3$ ground-truth actions captured at $20$ Hz, the dashed lines show the predicted angular positions for the Bobcat E10 arm's joints. Note that the terms $\tilde{\bm{x}}_{t+j}$, $j\in\{1,..., H-1\}, H=10^3$, are sequentially constructed by updating the components of $\tilde{\bm{x}}_t$ with ground-truth actions and model predictions.}
    \label{fig:open_loop_predictions}
\end{figure*}

This data, which amounts to a total of approx. 76 min of teleoperation, is used to learn a model for the hydraulic arm's actuators response, as will be discussed in the following section.

\subsection{Training and evaluation of models for the arm dynamics}
\label{subsec:dynamic_model_training_eval}

The teleoperation dataset $\mathcal{D}$ contains temporally ordered tuples $(\bm{q}_t, \dot{\bm{q}}_t, \bm{a}_t)$. In principle, this dataset can be constituted by the aggregation of several trajectories (with a length of at least $H+1$ elements each, plus the data to construct $\tilde{\bm{x}}_t$), as long as these are managed separately when sampling short sequences ${\tau_i=(\bm{q}_{t_i}, \dot{\bm{q}}_{t_i}, \bm{a}_{t_i}, ..., \bm{q}_{t_{i}+H}, \dot{\bm{q}}_{t_{i}+H}, \bm{a}_{t_{i}+H})}$, for a given initial time step $t_{i}$. In practice, the dataset we use for training is composed by the three trajectories depicted in Fig.~\ref{fig:db_trajectories}, and is divided into training and validation subsets. The training subset is constructed by using $90\%$ of each trajectory, starting from the first sample, whereas the evaluation subset is constructed using the remaining portion of the trajectories.

Following the method described in Section~\ref{sec:dynamic_model}, to learn the parameters $\bm{\theta}$ of a given dynamic model $f_{\bm{\theta}}$, we minimize the loss function defined in Eq.~\eqref{eq:dmodel_loss_function}, which is computed by sampling the trajectories $\tau_i$ from $\mathcal{D}$. Note that evaluating $\mathcal{L}(\bm{\theta})$ requires setting a temporal horizon $H$, the number of temporal lags on angular positions/velocities and actions used to construct $\tilde{\bm{x}}_t$, which is given by $k$, and an architecture for an MLP or KAN to represent $f_{\bm{\theta}}$. Moreover, as discussed in Section~\ref{sec:dynamic_model}, we can choose to predict angular position residuals, $\Delta\bm{q}$, or angular velocity residuals, $\Delta\dot{\bm{q}}$.

To manage the aforementioned possibilities, for each prediction type and for each type of neural network (MLP/KAN), we search for suitable hyperparameters through the optimization of the metric $\mathcal{L}(\bm{\theta})|^\text{eval}_{H=80}$, which has the same definition as the loss function used for training (see Eq.~\eqref{eq:dmodel_loss_function}), but is evaluated using $100$ short trajectories sampled from the evaluation subset over a fixed time horizon, $H=80$. This process is done using the Optuna library~\citep{akiba2019optuna}. More details on this optimization process are documented in Appendix~\ref{appendix:dynamic_models_training_details}.

The above produces a ranking of models with respect to the evaluation metric $\mathcal{L}(\bm{\theta})|^\text{eval}_{H=80}$. Since the top ranked models only differ by a slight margin with respect to $\mathcal{L}(\bm{\theta})|^\text{eval}_{H=80}$, we filter those with higher lags $k$ and prefer smaller models due to their faster inference time (which is specially relevant for the policy obtained using iCEM). Under these criteria, the selected models are characterized in Table~\ref{tab:best_dynamic_models}.

\begin{table}[]
    \centering
    \caption{Selected parameterizations and training hyperparameters for the dynamic models.}
    \begin{tabular*}{\linewidth}{c@{\extracolsep{\fill}}c@{\extracolsep{\fill}}c@{\extracolsep{\fill}}c@{\extracolsep{\fill}}c@{\extracolsep{\fill}}c@{\extracolsep{\fill}}c}
       \toprule
         \textbf{Pred.} & \textbf{Parameterization} & $\bm{k}$ & $\bm{H}$ & $\bm{lr}$& $\bm{n_B}$& $\bm{\mathcal{L}(\bm{\theta})|^\text{\textbf{eval}}_{H=80}}$ \\\midrule
          \multirow{2}{*}{$\Delta \bm{q}$} & $\text{MLP}(128, 128)$  &  $18$ & $8$ & $10^{-3}$& $512$&$5.586\cdot10^{-3}$\\
            & $\text{KAN}(32, 16, 32)$  & $18$ & $8$ & $10^{-5}$&$1024$& $5.563\cdot10^{-3}$\\ \midrule
          \multirow{2}{*}{$\Delta \dot{\bm{q}}$} & $\text{MLP}(512, 128)$  &  $18$ &$16$ &$10^{-3}$ &$256$& $5.675\cdot10^{-3}$\\
           & $\text{KAN}(32, 32, 16)$  & $18$ & $16$ & $10^{-4}$& $512$ & $5.555\cdot10^{-3}$\\
         \bottomrule
    \end{tabular*}
    \vspace{2pt}
    
    \footnotesize{$k$: temporal lags; $H$: training loss horizon; $lr$: learning rate; $n_B$: batch size.}
    \label{tab:best_dynamic_models}
\end{table}

Since these learned models are meant to serve as a surrogate for the real Bobcat E10 arm dynamics, their performance is further assessed in an open-loop prediction setting. Fig.~\ref{fig:open_loop_predictions} shows a ground-truth trajectory of configurations and normalized discrete commands sampled from the evaluation subset, alongside the configurations predicted by the learned dynamic models. The ground-truth trajectory consists of $10^3$ configuration-action pairs, which (as described in Section~\ref{subsec:data_collection}) are captured at $20$ Hz, so the open-loop predictions result in a rollout of $50$ seconds. Note that the models only have access to a first ground-truth construction of $\tilde{\bm{x}}_t$ and to the sequence of ground truth actions, therefore, prediction errors compound. 

The results obtained show that all the dynamic models can approximately capture the response of the actuators with low error (for short time horizons) in the evaluation subset. The dynamic models not being perfect is an expected result, given that their inputs lack relevant variables that influence the machine's responses, such as differential hydraulic pressures, the hydraulic fluid's temperature, and the motor's RPM. Moreover, the models are trained under a low data regime that does not contain samples from excitation signals purposefully designed for system identification, but only from the arm teleoperation.

Note that since the purpose of these models is to enable the synthesis of reaching controllers via RL or MPC, we want them not only to achieve low error in the teleoperation evaluation subset, but also to behave properly outside the training distribution. The above is crucial because we want the reaching controllers to perform appropriately for arbitrary initial and target poses sampled from the restricted workspace. Moreover, out-of-distribution (OOD) action sequences will likely be fed to the dynamic models during the RL training and the MPC optimization, so inconsistent dynamic model transitions for OOD inputs will result in policies with a high reality gap, which may render them useless.

Given the above, it is not clear whether there is an overall best dynamic model across those selected through hyperparameter optimization. Therefore, using each of them to synthesize policies seems a necessary experimental step, as only through the comparison of their respective policies it would be possible to know which dynamic model serves its purpose the best. This is done in the next section.

\subsection{Synthesis and evaluation of controllers for reaching}
\label{subsec:training_eval_brax}

Using the best learned dynamic models, we can synthesize policies for reaching. To do so, the learned dynamic models $f_{\bm{\theta}}$ are used as surrogates of the real dynamics of the mini-excavator arm. Given an action $\bm{a}_t$ and an initial history of angular positions/velocities and previous actions to construct $\tilde{\bm{x}}_t$, we can get a next configuration for the hydraulic arm. Given a new control signal $\bm{a}_{t+1}$ and the predicted configuration to construct $\tilde{\bm{x}}_{t+1}$, and repeating the above procedure, we can obtain a sequence $(\tilde{\bm{x}}_t, \bm{a}_t, ..., \tilde{\bm{x}}_{T}, \bm{a}_{T})$ for a given time horizon~$T$. If we refer to Section~\ref{subsubsec:modeling}, we can see that this sequence can then be used to compute a sequence of observations, actions, and rewards, $(\bm{o}_t, \bm{a}_t, r_t, ... \bm{o}_T, \bm{a}_T, r_T)$, which we can leverage to obtain policies via RL or MPC (see Sections~\ref{subsubsec:rl_controllers} and~\ref{subsubsec:mpc_controllers}). 

It is important to note that the selection of a given dynamic model to train an RL-based policy conditions the policy's observations, because observations are constructed using $(\tilde{\bm{x}}_t, \bm{a}_t)$ tuples, and the definition of $\tilde{\bm{x}}_t$ varies depending on the dynamic model's temporal lags and whether it predicts angular position or angular velocity residuals. However, since all selected dynamic models share the same number of temporal lags $k$ (see Table~\ref{tab:best_dynamic_models}), the main difference between the observations used as input for the RL-based policies is simply whether they are constructed using normalized angular positions or angular velocity estimations (see Eqs.~\eqref{eq:x_t_pos} and~\eqref{eq:x_t_vel}). 

\begin{figure}
    \centering
    \includegraphics[width=0.6\linewidth]{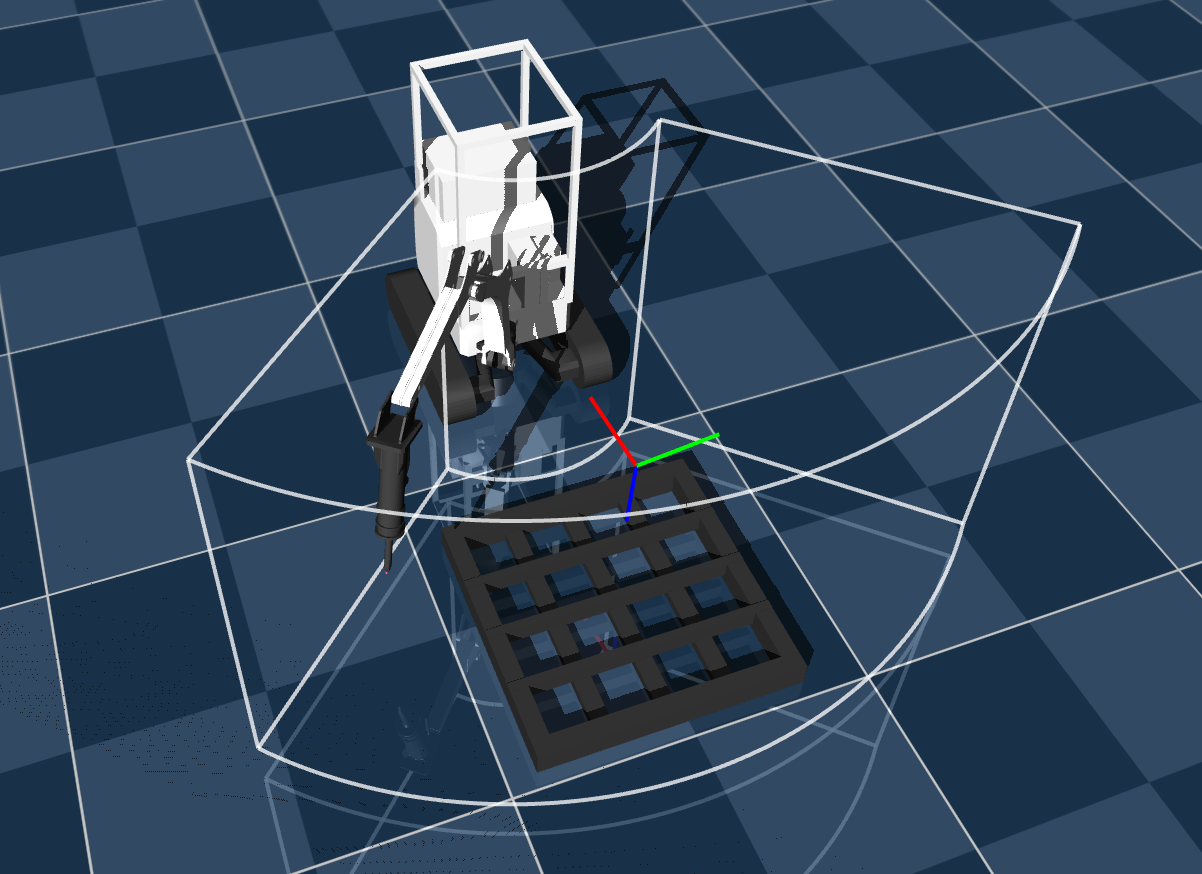}
    \caption{Training environment renderization in the MuJoCo visualizer, with markers to represent the restricted workspace $\mathcal{W^+}$ and a randomly chosen target end-effector pose.}
    \label{fig:mujoco_rendered_env}
\end{figure}

To implement the above, we use the JAX-based computational pipeline described in Section~\ref{subsec:computational_pipeline}. Specifically, we implement a Brax-compliant environment that specifies the observations, actions, reward function, and episodic settings of the problem modeling, as described in Section~\ref{subsubsec:modeling}, and use the Bobcat E10 robot description illustrated in Fig.~\ref{fig:bobcat_e10_kinematics} to manage aspects such as the forward kinematics computation. This allows using the PPO implementation provided by Brax, and to easily parallelize the execution of rollouts based on the learned dynamic model predictions, which is crucial to achieve real-time control using MPC. The resulting environment can be rendered using the MuJoCo visualizer, as shown in Fig.~\ref{fig:mujoco_rendered_env}.

Using this environment, we can train RL-based policies using PPO for $400\text{M}$ steps, which only takes about $6$~min. using a mid-range computer. Similarly, we can instantiate iCEM-based controllers that will provide action sequences that aim at maximizing cumulative rewards over short horizons (said rewards being obtained using the same reward function used for RL).

To evaluate the performance of the policies, we consider the following metrics:

\begin{table*}[]
    \sisetup{separate-uncertainty=true, table-align-uncertainty=true}
    \centering
    \caption{Evaluation results obtained in the Brax-based training environment. The performance metrics are computed across $500$ independent reaching trials (episodes). The reported average and standard deviation per metric is computed across five independently synthesized controller per policy type. }
    \label{tab:training_metrics}
    \begin{tabular*}{\linewidth}{c@{\extracolsep{\fill}}c@{\extracolsep{\fill}}S[table-format=-3.2(5)]@{\extracolsep{\fill}}c@{\extracolsep{\fill}}c@{\extracolsep{\fill}}c@{\extracolsep{\fill}}c@{\extracolsep{\fill}}S[table-format=-1.2(3)]}
    \toprule
       \textbf{Algo.} & \textbf{Dyn. model} & {\textbf{Avg. return}} & \textbf{SR}$\bm{(0.02, 0.02)}$ &  \textbf{SR}$\bm{(0.04, 0.04)}$ & \textbf{OO}$\bm{\mathcal{W}}$ & \textbf{OO}$\bm{Z}$ & {\textbf{Min.} $\bm{z}$ \textbf{[m]}} \\ \midrule
        \multirow{4}{*}{PPO} & MLP $\Delta \bm{q}$ & 644.45 \pm 10.31& $0.97\pm0.03$ &$1.00 \pm 0.00$& $0.13 \pm 0.01$ & $0.08 \pm 0.01$& -0.22 \pm 0.28\\
         & MLP $\Delta \dot{\bm{q}}$ & 683.36 \pm 16.94& $0.99 \pm 0.01$& $1.00 \pm 0.00$&  $0.15 \pm 0.01$ & $0.11 \pm 0.01$ & -0.10 \pm 0.08 \\
         & KAN $\Delta \bm{q}$ & 458.85 \pm 17.13 & $0.91 \pm 0.02$ & $0.98\pm 0.01$& $0.15 \pm 0.02$ & $0.10 \pm 0.02$ & -1.17 \pm 0.04 \\
         & \, KAN $\Delta \dot{\bm{q}}$$^*$ & -220.78 \pm 604.06 & $0.26 \pm 0.21 $& $0.42 \pm 0.35$ & $0.36 \pm 0.25$ & $0.24 \pm 0.25$ & -0.71 \pm 0.34 \\ \midrule
        iCEM & \multirow{3}{*}{MLP $\Delta \dot{\bm{q}}$}& 15.96 \pm 5.95 & $0.20 \pm 0.01$ & $0.28 \pm 0.01$ & $0.03 \pm 0.01$& $0.02 \pm 0.01$ &  0.14 \pm 0.01\\
        \, iCEM$^{\dagger}$ & & -348.59 \pm  1.10 & $0.66 \pm 0.01$ & $0.78 \pm 0.00$& $0.02 \pm 0.00$& $0.01 \pm 0.00$ & 0.13 \pm 0.02 \\ 
        \, iCEM$^{\ddagger}$ & &-148.48 \pm 0.31& $0.81 \pm 0.01$& $0.91 \pm 0.01$ & $0.01 \pm 0.00$&   $0.00 \pm 0.00$ & 0.14 \pm 0.01 \\ 
         \bottomrule
    \end{tabular*}
    \vspace{2pt}
    
    \footnotesize{\raggedright $^*$Learning rate set to $0.8\cdot10^{-4}$ and evaluated at $800$M training steps. $^{\dagger}$$r^{\epsilon}_t$ set to zero.
    $^{\ddagger}$$r^{\epsilon}_t$ and $r_t^{\mathcal{X}}$set to zero.\hfill}
   
\end{table*}

\begin{itemize}

    \item Avg. return: Undiscounted sum of instantaneous rewards obtained up until $T_{\text{max}}=500$ steps pass, averaged across evaluation episodes. 

    \item Success rate (conditioned on precision requirements): Given a threshold for the Euclidean distance between the current and target end-effector poses, $\varepsilon_{\bm{p}}$, and a threshold for the absolute difference between their pitch angles, $\varepsilon_{\alpha}$, we consider the reaching task as successful if $b_{\text{SR}}$, defined by Eq.~\eqref{eq:b_sr}, equals one at least once during a given episode.
    \begin{equation}
        b^t_{\text{SR}} =  \bigl\llbracket d(\bm{p}^{\text{target}},\bm{p}_t^{\text{eef}})<\varepsilon_{\bm{p}})\bigr \rrbracket \cdot \llbracket | \alpha_{\text{pitch}} - \alpha_{\text{pitch}}^{\text{target}}|<\varepsilon_\alpha \rrbracket
        \label{eq:b_sr}
    \end{equation}

    Thus, the success rate (SR) of a reaching policy is computed as $\text{SR}(\varepsilon_{\bm{p}}, \varepsilon_\alpha):=\frac{1}{N}\sum_{i=1}^N \llbracket\sum_{t=1}^{T_{\text{max}}}  b_{\text{SR}}^{t,(i)}\geq1\rrbracket$, where the dependence on threshold selection is highlighted, and the metric is computed for $N$ episodes, each of them with a time horizon $T_\text{max}$.

    \item Rate at which the end-effector leaves the restricted workspace $\mathcal{W}^+$ (OO$\mathcal{W}$): Consider the variable $b_{\mathcal{W}}$ defined in Eq.~\eqref{eq:b_w}, which equals one if the end effector is within the restricted workspace.
    \begin{equation}
        b^t_{\mathcal{W}}=\llbracket(\mathcal{ X}_t^{\text{eef}}\in\mathcal{W}^ {+})\rrbracket
        \label{eq:b_w}
    \end{equation}

    Similarly to the success rate, the out-of-workspace rate is  defined as $\text{OO}\mathcal{W}:=\frac{1}{N}\sum_{i=1}^N\llbracket\sum_{t=1}^{T_\text{max}}(1-b^{t,(i)}_{\mathcal{W}})\geq1\rrbracket$ for $N$ evaluation episodes, where each of them has a time horizon~$T_\text{max}$.

    \item Rate at which the end-effector goes below the lower $z$-axis boundary of $\mathcal{W}^+$ (OO$Z$): Defined analogous to OO$\mathcal{W}$. Note that OO$Z$ $\leq$ OO$\mathcal{W}$.

    \item Minimum $z$-axis component of the end-effector position across all the evaluated trajectories: This metrics measures the worst possible end-effector breach through the lower $z$-axis plane that defines the restricted workspace.

\end{itemize}

With the above, we train RL-based policies for each best learned dynamic model (which are characterized in Table~\ref{tab:best_dynamic_models}). We consider five independent training trials per policy and evaluate them over a set of $N=500$ episodes, each of them with fixed but different episodic conditions. For MPC-based policies, we do the same, but only using the MLP $\Delta\dot{\bm{q}}$ model, since it allows synthesizing the best RL-based controllers, as it will be explained later. Moreover, for all the policies we set a control frequency of $20$ Hz (matching the time discretization of the dynamic models) and $4$ action repeats. In addition, all the RL-based policies are evaluated after $270$M training steps, except for the PPO KAN$\Delta\dot{\bm{q}}$ whose learning rate and training steps were further tuned due to worse performance being obtained when training under the same conditions than the rest of the controllers. The results obtained are presented in Table~\ref{tab:training_metrics} and all the training details and hyperparameters used to obtain policies are documented in Appendix~\ref{appendix:policy_learning_details}. 

From training RL-based policies using PPO, we verified that using different dynamic models has an impact on the final controller's performance. The first indicator of this difference lies in the cumulative reward obtained at the end of the training process, where the policies trained using the best KAN $\Delta \dot{\bm{q}}$ model achieve far lower avg. returns than the rest, which motivates filtering them from further analysis. We can also note a difference in average return (albeit less pronounced) between the policies trained using the KAN $\Delta \bm{q}$ model and those trained using MLPs as surrogate dynamic models, the latter achieving better overall performances. It is also observed that the breaches through the lower plane of the restricted workspace are much less pronounced for the PPO-MLP $\Delta \dot{\bm{q}}$ model, as the ``min. $z$'' metric is lower for this controller when compared to the rest of the RL-based policies.

From training MPC-based policies, we note that optimizing slight variations of the reward function (by setting $r_t^{\epsilon}=0$ and $r_t^{\epsilon}=r_t^{\mathcal{X}}=0$) improves performance. This result can be explained considering that $r_t^{\epsilon}$ is a positive scalar only given to the agent when certain conditions are fulfilled, which makes the reward function highly non linear and may make its optimization over short horizons more susceptible to convergence to local optima. Moreover, also setting $r_t^{\mathcal{X}}=0$ further improves results for iCEM, which can be explained considering the reward ablation documented in Appendix~\ref{appendix:policy_learning_details} for an RL based policy (using the same dynamic model used by iCEM), which suggests that $r_t^{\bm{q}}$ and $r_t^{\mathcal{X}}$ may be conflicting terms when optimized simultaneously. When comparing the overall performance of controllers based on RL and MPC, we can see that although RL policies in general achieve a higher success rate, iCEM controllers are better at avoiding breaching the restricted workspace boundaries. This result may be due to iCEM discarding the action sequences that would result in such breaches during optimization, and PPO policies, on the other hand, affording the cost associated with leaving $\mathcal{W}^{+}$ whenever a higher return (in the long run) can be achieved.

The results motivate studying the behavior of the best policies obtained in the real world. To do so, in the next section we conduct experiments to measure the performance of all the RL-based policies (except the PPO-KAN $\Delta\dot{\bm{q}}$ policies) and the best iCEM controller when undergoing a Sim2Sim and a Sim2Real transfer.

\subsection{Sim2Sim and Sim2Real transfer of the learned controllers}
\label{subsec:sim2sim_andsim2real}

We deploy the synthesized policies in the real-world by leveraging a ROS-based computational pipeline. Nevertheless, a pre-deployment stage using the Gazebo simulator was utilized to test the different system components prior to using them to control the real mini-excavator. In both the pre-deployment (Sim2Sim) and real-world deployment (Sim2Real) stages, the reaching controllers are encapsulated in action servers that, once queried with a target pose, will execute actions (according to an RL or MPC-based policy) until successfully reaching the target or preemptively stopping when an undesired behavior is detected. We must note, however, that to isolate the performance of the policies, all safety measures are disabled during the Sim2Sim and Sim2Real transfer experiments. The computational pipeline that allows implementing the pre-deployment and real-world deployment stages is described in Appendix~\ref{appendix:ros_based_pipelines}. In what follows, the performed experiments and their results are documented.

\subsubsection{Sim2Sim transfer}
\label{subsec:sim2sim}

A first pre-deployment experiment is conducted using Gazebo with the reaching controllers encapsulated in ROS-based action servers. When evaluating a given policy, the learned dynamic model used to synthesize it is encapsulated in a module that allows bypassing the Gazebo simulator physics. This is done by using the inferences of these dynamic models, conditioned on previous joint positions/velocities and actions, to set the hydraulic arm configuration directly, similar to what is done in the Brax-based environment (see Appendix~\ref{appendix:ros_based_pipelines}). 

The Sim2Sim transfer experiments use the exact same initial and target end-effector poses set during the experiments conducted in the Brax-based environment, for the same $500$ distinct episodes (see Section~\ref{subsec:training_eval_brax}). Moreover, as in the Brax-based environment evaluation, the execution of the action servers for a given target pose is only terminated by timeout (after $500$ time steps). In this case, however, only a single policy is evaluated (which corresponds to the first of the five synthesized controllers for each policy type). Under these conditions, this Sim2Sim transfer study allows measuring the effects of time delays due to the communication of messages to construct observations and set control commands in an isolated manner (as the learned dynamic models dictate the response of the impact hammer to actions). It is worth mentioning that such delays are not present in the Brax-based training environment, but exist during real world deployment. 

To get a fine-grained performance measure of the controllers when evaluated under this new setting, we construct success rate matrices by computing the SR metric with thresholds $(\varepsilon_{\bm{p}}, \varepsilon_\alpha)\in\{0.02, 0.04,..., 0.34\}\times\{0.02, 0.04, ..., 0.16\}$, which can be visualized in Appendix~\ref{appendix:sim2sim}. Table~\ref{tab:sim2sim} shows the performance metrics obtained, highlighting the success rates obtained at $(\epsilon_{\bm{p}},\epsilon_{\alpha})=(0.02,0.02)$, and $(\epsilon_{\bm{p}},\epsilon_{\alpha})=(0.12,0.08)$. 

\begin{table}[]
    \centering
    \caption{Sim2Sim transfer evaluation results obtained across $500$ independent reaching trials.}
    \label{tab:sim2sim}
    \begin{tabular*}{\linewidth}{l@{\extracolsep{\fill}}c@{\extracolsep{\fill}}c@{\extracolsep{\fill}}c@{\extracolsep{\fill}}c@{\extracolsep{\fill}}c@{\extracolsep{\fill}}c@{\extracolsep{\fill}}S[table-format=-1.3]}
    \cmidrule[0.08em]{2-8}
         &\textbf{Algo.} & \textbf{Dyn. model} & \textbf{SR}$^a$ & \textbf{SR}$^b$ & \textbf{OO}$\bm{\mathcal{W}}$ & \textbf{OO}$\bm{Z}$ & {\textbf{Min.} $\bm{z}$ \textbf{[m]}}  \\ %
         \midrule
         \multirow{4}{*}{\rotatebox[origin=c]{90}{\textbf{Brax}}}
         &\multirow{3}{*}{PPO} & MLP $\Delta \bm{q}$ &$0.99$&$1.00$ &$0.124$ & $0.068$ & -0.066  \\
         && MLP $\Delta \dot{\bm{q}}$ &$0.98$& $1.00$ &$0.130$ & $0.092$ & -0.103 \\
         && KAN $\Delta \bm{q}$ &$0.92$ &$1.00$ &$0.136$ & $0.088$ & -1.184  \\
         &iCEM$^{\ddagger}$ & MLP $\Delta \dot{\bm{q}}$ & $0.81$&$0.98$ &$0.012$ & $0.004$ & 0.124 \\
         \midrule
         \multirow{4}{*}{\rotatebox[origin=c]{90}{\textbf{Gazebo}}}
         &\multirow{3}{*}{PPO} & MLP $\Delta \bm{q}$ & $0.02$ &$0.93$ &$0.124$ & $0.066$ & -0.004  \\
         && MLP $\Delta \dot{\bm{q}}$ & $0.11$ &$1.00$ &$0.118$ & $0.068$ & 0.000 \\
         && KAN $\Delta \bm{q}$ & $0.12$ &$0.90$& $0.142$ & $0.086$ & -0.574  \\
         &iCEM$^{\ddagger}$ & MLP $\Delta \dot{\bm{q}}$ &$0.74$ &$0.98$ &$0.122$ & $0.068$ & 0.107  \\
         \bottomrule
    \end{tabular*}
    \vspace{2pt}
    
    \footnotesize{\raggedright$^{a}$SR$(0.02, 0.02)$. $^b$SR$(0.12, 0.08)$. $^{\ddagger}$$r^{\epsilon}_t$ and $r_t^{\mathcal{X}}$set to zero.\hfill}
\end{table}

The results show a noticeable performance drop for the success rates at high precision requirements for the RL-based policies, while this is not the case for the iCEM controller. We attribute this to two complementary causes: the differences in the reward function optimized by RL-based policies and the iCEM controller, and the asynchronous nature of the Gazebo-ROS computational pipeline, which introduces delays that result in perturbations on the environment state transitions. Firstly, PPO policies are trained using the full definition of $r_t$ (see Eq.~\eqref{eq:reward_function}), while iCEM policies consider a slight variation of this function, where the terms $r_t^{\mathcal{X}}$ and $r_t^{\epsilon}$ are set to zero. One can think of $r_t^{\epsilon}$ as a bonus term encouraging the end-effector to stabilize on the target pose, since it is only non-zero if precision requirements over the end-effector pose and the arm configuration are fulfilled (see Eqs.~\eqref{eq:b_pose}--\eqref{eq:b_q}). This makes oscillations around the target pose less pronounced once it has been reached at a certain level of precision, however, when deploying the controllers in the Gazebo simulator, the dynamic models, the reaching controller, and the simulator itself, all run asynchronously, and that induces time delays that are not present when training and evaluating the policies in Brax. These perturbations harm the performance of RL-based policies at low success rate thresholds. In contrast, the iCEM controller tends to oscillate around the target (since it does not optimize for $r_t^{\epsilon}$), which makes it more likely to eventually approach the target pose (during the whole execution of a given episode), regardless of the aforementioned perturbations. We must note, however, that iCEM is also affected by the asynchronous nature of Gazebo, as the OO$\mathcal{W}$ and OO$Z$ metrics are higher in this environment compared to Brax, however, the Min. $z$ coordinate of its end-effector during all the experiments is still higher than those of RL-based policies. Despite these observations, when deployed in Gazebo, one can see that the PPO MLP $\Delta\dot{\bm{q}}$ controller nearly matches the SR of the iCEM controller at thresholds $(\epsilon_{\bm{p}},\epsilon_{\alpha})=(0.06,0.06)$, and surpasses it at $(\epsilon_{\bm{p}},\epsilon_{\alpha})=(0.08,0.04)$ and higher (see Fig.~\ref{fig:sr_matrices_brax_gazebo} in Appendix~\ref{appendix:sim2sim}). Moreover, it matches or surpasses the precision levels of the rest of RL-based policies.

\subsubsection{Sim2Real transfer}

The policies are deployed in the real Bobcat E10 mini-excavator using the same ROS-based action servers used for the Sim2Sim transfer experiments. Again, the execution of the action servers is only stopped by a timeout after $500$ time steps pass. Since in this case it is not possible to easily set random initial configurations for each reaching episode, given an initial configuration for the hydraulic arm, the evaluation experiments consist of attempting to reach $100$ poses, which are sequentially fed to the action servers. Given that episodes only finish by time-out, this translates in the target end effector pose changing every $500$ time-steps, making the initial condition for episodes (except for the first one) dependent on the target pose of the previous episode and the performance of the policy itself. Moreover, the $100$ target end-effector poses coincide with the first $100$ targets that are used for the evaluation of policies in simulation, both for the Brax-based environment and when using the Gazebo-ROS pipeline. 

\begin{figure*}
    \centering
    \subfloat[][]{\includegraphics[width=\linewidth]{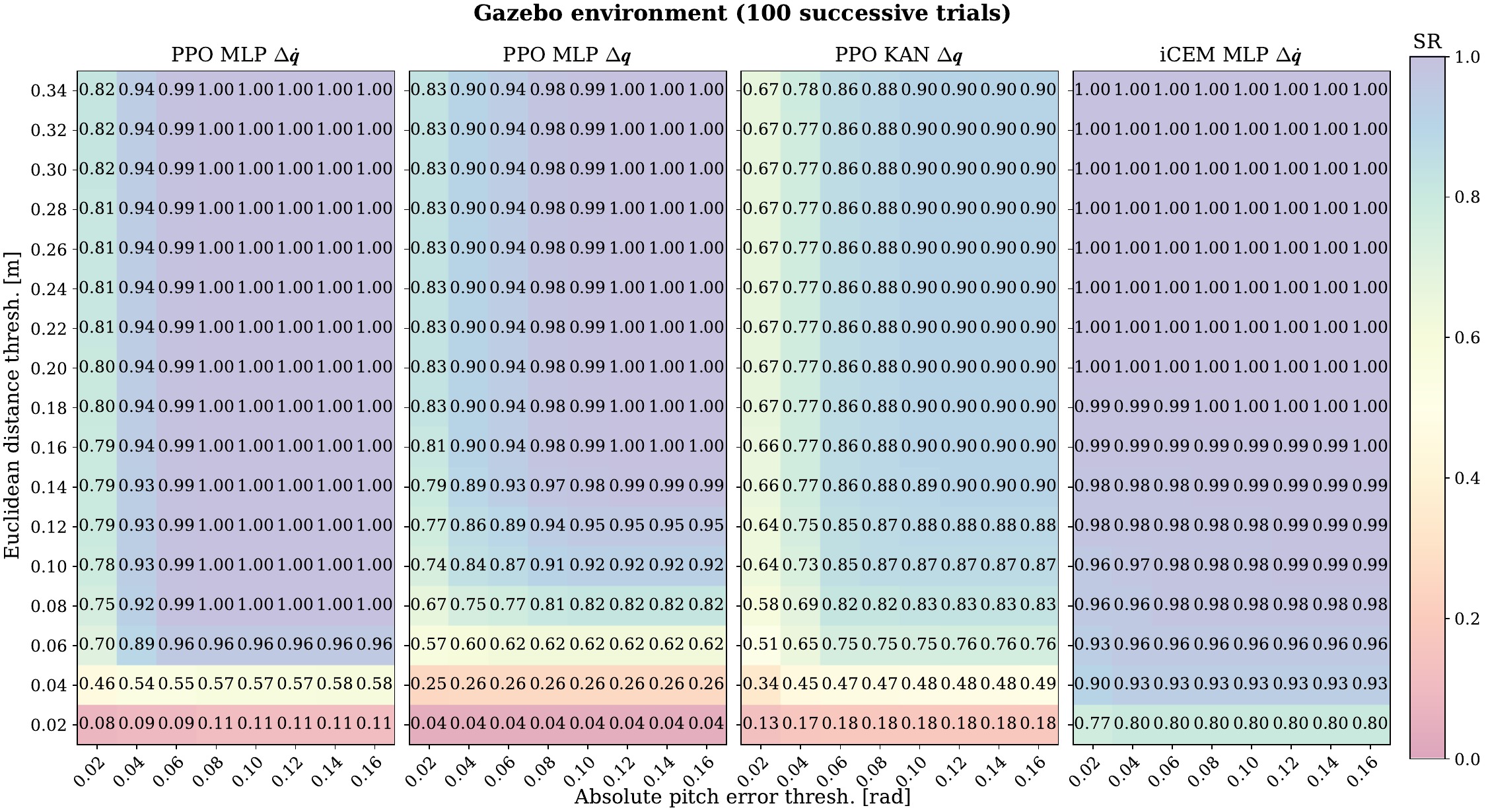}}\\ 
    \subfloat[][]{\includegraphics[width=\linewidth]{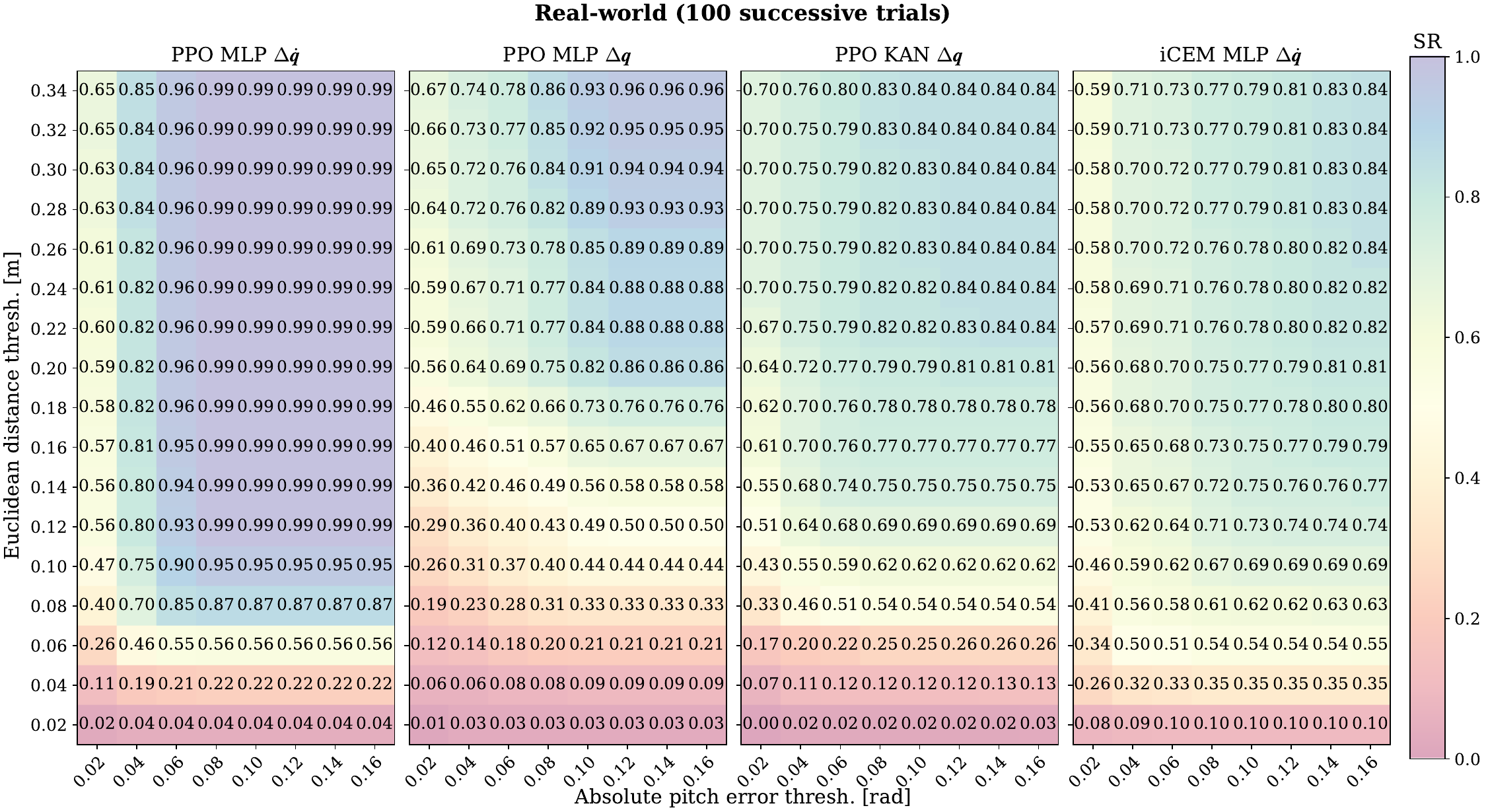}\label{fig:sr_matrices}} 
    \caption{Success rate matrices obtained when evaluating the reaching policies in (a) the Gazebo simulator and (b) the real-world, in both cases, by encapsulating the policies in action servers.}
    \label{fig:sr_matrices_real_world}
\end{figure*}

\begin{table}[]
    \centering
    \caption{Sim2Real transfer evaluation results obtained across $100$ successive reaching trials.}
    \label{tab:sim2real}
    \begin{tabular*}{\linewidth}{l@{\extracolsep{\fill}}c@{\extracolsep{\fill}}c@{\extracolsep{\fill}}c@{\extracolsep{\fill}}c@{\extracolsep{\fill}}c@{\extracolsep{\fill}}c@{\extracolsep{\fill}}c}
    \cmidrule[0.08em]{2-8}
         &\textbf{Algo.} & \textbf{Dyn. model} & \textbf{SR}$^a$ & \textbf{SR}$^b$ & \textbf{OO}$\bm{\mathcal{W}}$ & \textbf{OO}$\bm{Z}$ & \textbf{Min.} $\bm{z}$ \textbf{[m]} \\
         \midrule
         \multirow{4}{*}{\rotatebox[origin=c]{90}{\textbf{Gazebo}}}
         &\multirow{3}{*}{PPO} & MLP $\Delta \bm{q}$ &$0.04$& $0.94$ &$0.040$ & $0.010$ & $0.091$  \\
         && MLP $\Delta \dot{\bm{q}}$ &$0.08$& $1.00$&$0.090$ & $0.020$ & $0.005$  \\
         && KAN $\Delta \bm{q}$ &$0.13$& $0.87$&$0.060$ & $0.030$ & $0.000$  \\
         &iCEM$^{\ddagger}$ & MLP $\Delta \dot{\bm{q}}$ &$0.77$&$0.98$ &$0.070$ & $0.020$ & $0.151$  \\
         \midrule
         \multirow{4}{*}{\rotatebox[origin=c]{90}{\textbf{R. World}}}
         &\multirow{3}{*}{PPO} & MLP $\Delta \bm{q}$ &$0.01$& $0.43$ &$0.010$ & $0.010$ & $0.154$  \\
         && MLP $\Delta \dot{\bm{q}}$ &$0.02$&$0.99$ &$0.070$ & $0.040$ & $0.050$  \\
         && KAN $\Delta \bm{q}$ &$0.00$&$0.69$ &$0.020$ & $0.020$ & $0.099$  \\
         &iCEM$^{\ddagger}$ & MLP $\Delta \dot{\bm{q}}$ &$0.08$&$0.71$ & $0.200$ & $0.050$ & $0.120$  \\
         \bottomrule
    \end{tabular*}
    \vspace{2pt}
    
    \footnotesize{\raggedright$^{a}$SR$(0.02, 0.02)$. $^b$SR$(0.12, 0.08)$. $^{\ddagger}$$r^{\epsilon}_t$ and $r_t^{\mathcal{X}}$set to zero.\hfill}
\end{table}

\begin{figure*}
    \centering
    \includegraphics[width=\linewidth]{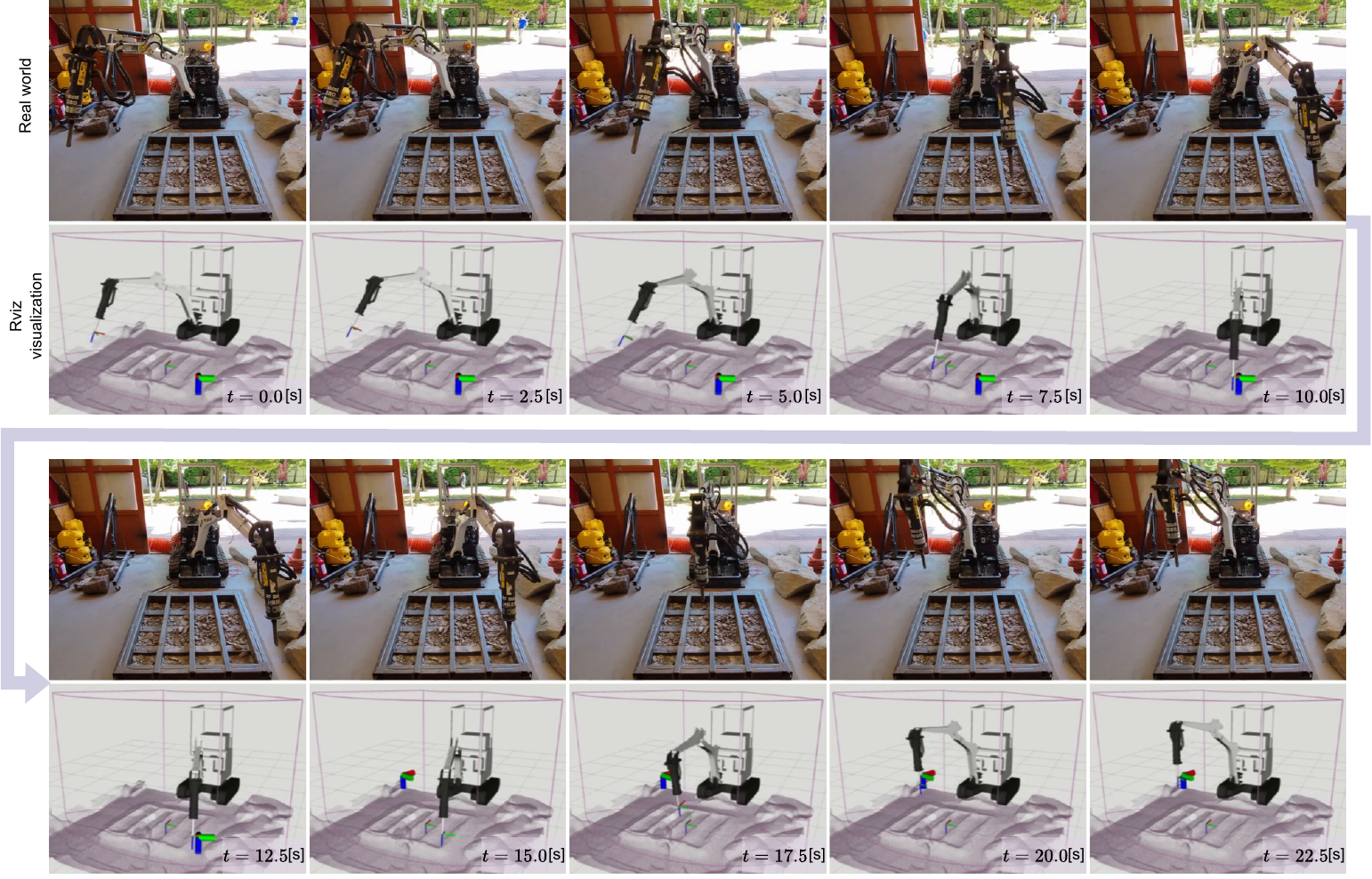}
    \caption{Real-world PPO-MLP $\Delta\dot{\bm{q}}$ policy execution for two sequentially queried target poses.}
    \label{fig:pose_seq_real_world}
\end{figure*}
 
The execution of these experiments for policies trained using PPO with dynamic models MLP~$\Delta \dot{\bm{q}}$, MLP~$\Delta \bm{q}$ and KAN~$\Delta \bm{q}$, and for the iCEM controller alongside the MLP~$\Delta \dot{\bm{q}}$ model results in the success rate matrices shown in Fig.~\ref{fig:sr_matrices_real_world} and in the performance metrics summarized in Table~\ref{tab:sim2real}. Moreover, these experiments translate into approximately $50$ minutes of autonomous operation per evaluated controller, whose footage, for a qualitative evaluation of performance, can be seen in the following link: \url{https://youtu.be/lshZwPQMlFg}.

The results obtained show that RL-based controllers display a detriment on their success rate at low Euclidean distances and pitch angle error thresholds when comparing their performance in the real world and in the Gazebo-ROS pipeline. The reaching controller that is the least affected by the Sim2Real transfer and also achieves better overall performance remains the PPO-MLP $\Delta \dot{\bm{q}}$ policy, achieving a near perfect success rate for the sequential queried target poses $100$ at thresholds $(\epsilon_{\bm{p}}, \epsilon_{\alpha})=(0.12, 0.08)$. Moreover, contrary to what was observed in the Sim2Sim experiments for the iCEM controller, in the real world it drastically dropped its success rate for all thresholds while also achieving higher OO$\mathcal{W}$ and OO$Z$ values.

We note that a common failure mode observed for the PPO KAN $\Delta \bm{q}$ and the iCEM controller during real world deployment was that sometimes they were unable to generate commands that resulted in an actual displacement of the hydraulic arm actuators. This happened whenever the mini-excavator's arm reached certain configurations and specific target poses were being queried. In such situations, the policies would often output commands set to zero for all joints or would rapidly alternate the values of certain action components, which, given the hydraulic coupling and high actuation delays of the arm's hydraulic cylinders, resulted in no movement, regardless of the enforced action regularization via the reward function definition, and the action repeats used to synthesize and deploy the controllers. This behavior may be explained due to the learned dynamic models being imperfect and producing misleading transitions in certain regions of the state space, which translates into inadequate control strategies for certain situations and, therefore, poor performance when facing those situations in the real world.

The fact that the PPO MLP $\Delta \dot{\bm{q}}$ policy achieves good performance and does not display the failure modes that iCEM shows (although both depend on the same dynamic model), may be due to the PPO MLP $\Delta \dot{\bm{q}}$ policy, since parameterized as an MLP itself, has the capacity to ``average'' its behavior across poorly represented regions of the state space; moreover, given that its observations are mostly constructed using velocity estimations, it is likely that this policy puts less importance on the instantaneous configuration of the arm, which may help generalization.

Finally, in addition to the Sim2Real transfer experiment documented above, we further tested the PPO MLP $\Delta \dot{\bm{q}}$ policy, setting fixed euclidean distance and absolute pitch error thresholds to $\epsilon_{\bm{p}}=0.12$ [m] and $\epsilon_{\alpha}=0.08$ [rad] to decide whether a target pose was reached, so as to assess its performance in a setting more close to the practical application of the controller. The result of these experiments is qualitatively presented in Fig.~\ref{fig:pose_seq_real_world} for two sequentially queried target poses, and the corresponding video with further trials can be found in \url{https://youtu.be/e-7tDhZ4ZgA}.

\section{Conclusion}
\label{sec:conclusion}

In this work, we presented a practical, data-driven methodology to automate hydraulic impact hammers such as those used for secondary reduction in mining. This challenge is addressed considering operational constraints that may be present in real mining operations, such as unobserved state variables, and the requirement of relying on a discrete control interface. In particular, a methodology for learning dynamic models using teleoperation data and then exploiting said models to obtain policies for the task of reaching target end-effector poses is presented and validated. Said validation is performed both in simulation and the real world, using a Bobcat E10 mini-excavator with an impact hammer attached as end-effector.

The results obtained are promising given that the best obtained reaching policy is able to consistently reach target poses with relatively high precision in the real world, regardless of the mini-excavator arm  being inherently challenging to control due to partially observed states, actuation delays, and hydraulic coupling, and using discrete commands at the joint level to perform the reaching task.

However, the developed reaching controller has some drawbacks. One of its limitations is that it does not fully comply with the restrictions over the end-effector position given by the restricted workspace, and enforced by the reward term $r_t^{\mathcal{W}}$. This limitation could be alleviated by activating the safety measures included in the action server and using the control supervisor module included in the deployment computational pipeline (see Appendix~\ref{appendix:ros_based_pipelines}), since if the action server is stopped due to the end-effector leaving the workspace or due to the detection of oscillatory behaviors, a feedback signal reporting the unsuccessful completion of the reaching task could be sent to an hypothetical higher level system (e.g., a state machine) that in turn could trigger a recovery behavior. 

However, the most notable limitation of the reaching policies is that they are not informed by exteroceptive information, so it would not be able for them to avoid collisions if the constrained workspace in which the impact hammer operates ceases to be obstacle free. This would be troublesome for the deployment of these controller in a real mining operation, since frontal loaders and LHDs depositing material above ore passes' grills should be regarded as potential obstacles, regardless of additional safety measures. Moreover, the grill itself and the rocks remaining above it should also be regarded as obstacles, because grill geometries vary across  ore passes and moving from one target pose to another may induce an unintended collision with a rock. Therefore, developing a collision-aware reaching controller is left for future work.

Finally, the performance gap between iCEM and PPO-based policies in certain regions of the state space motivates investigating whether constraining the action search space to regions closer to the training distribution could improve MPC real-world performance.

\begin{figure*}[t!]
    \centering
    \includegraphics[width=\linewidth]{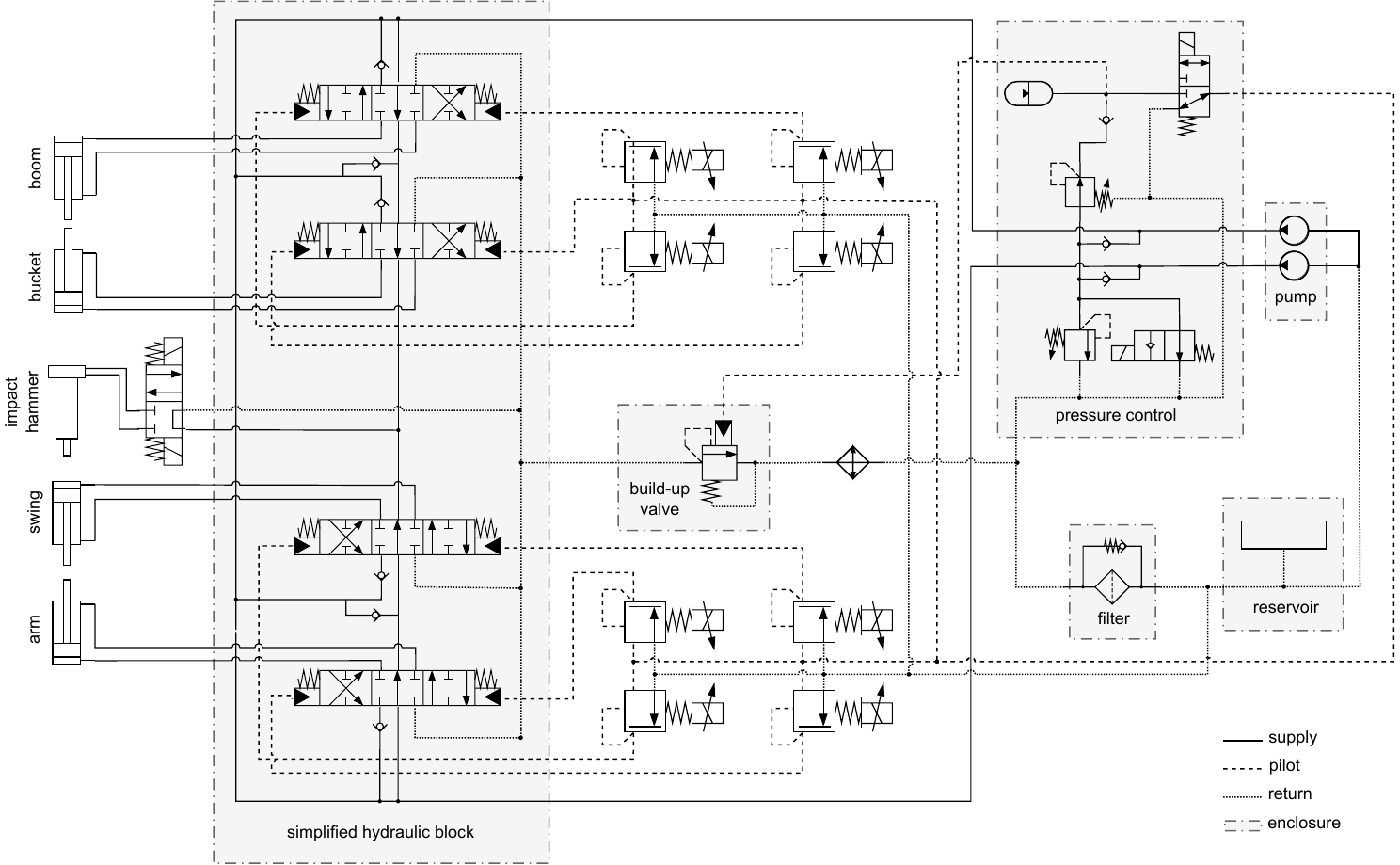}
    \caption{Simplified hydraulic circuit of the intervened Bobcat E10 mini-excavator (based on the original Bobcat E10 documentation and the modifications implemented on the machine).}
    \label{fig:bobcat_10_hydraulics}
\end{figure*}

\appendices

\section{Simplified Bobcat E10 hydraulic circuit}
\label{appendix:bobcat_e10_hydraulics}

Fig.~\ref{fig:bobcat_10_hydraulics} shows a simplified hydraulic circuit for the Bobcat E10 mini-excavator. From right to left, this diagram can be read as follows:
\begin{itemize}
    \item The supply line provided by the pumps goes through a pressure control sub-circuit that generates a pilot signal, and also directly feeds the (simplified) hydraulic block displayed in the diagram.
    \item The pilot signal feeds eight proportional electro-valves that are controlled via discrete current signals sent to them using an I/O module (see Fig.~\ref{fig:bobcat_e10_simplified_hydraulics}). The electro-valves' output goes to the valves within the hydraulic block shown in the diagram.
    \item The valves within the hydraulic block control the four actuators of the mini-excavator arm, that is, the hydraulic cylinders associated to the ``boom'', ``swing'', ``arm'' and ``bucket'' joints (see Fig.~\ref{fig:bobcat_e10_kinematics}). 
    \item The hydraulic block also feeds the supply line to an on-off electro-valve that actuates the impact hammer end-effector. Note that this on-off valve is also controlled by using the I/O module mentioned previously. 
\end{itemize}

Furthermore, this diagram shows some aspects of the Bobcat E10 mini-excavator that make controlling its arm challenging. Note, for instance, that controlling the arm's hydraulic cylinders depends on a pilot stage and a main hydraulic stage, which contributes to actuation delays. Also note that the valves within the hydraulic block share supply, which contributes to coupling across actuators. 

\section{ROS-based computational pipelines}
\label{appendix:ros_based_pipelines}
\begin{figure}
    \centering
    \includegraphics[width=\linewidth]{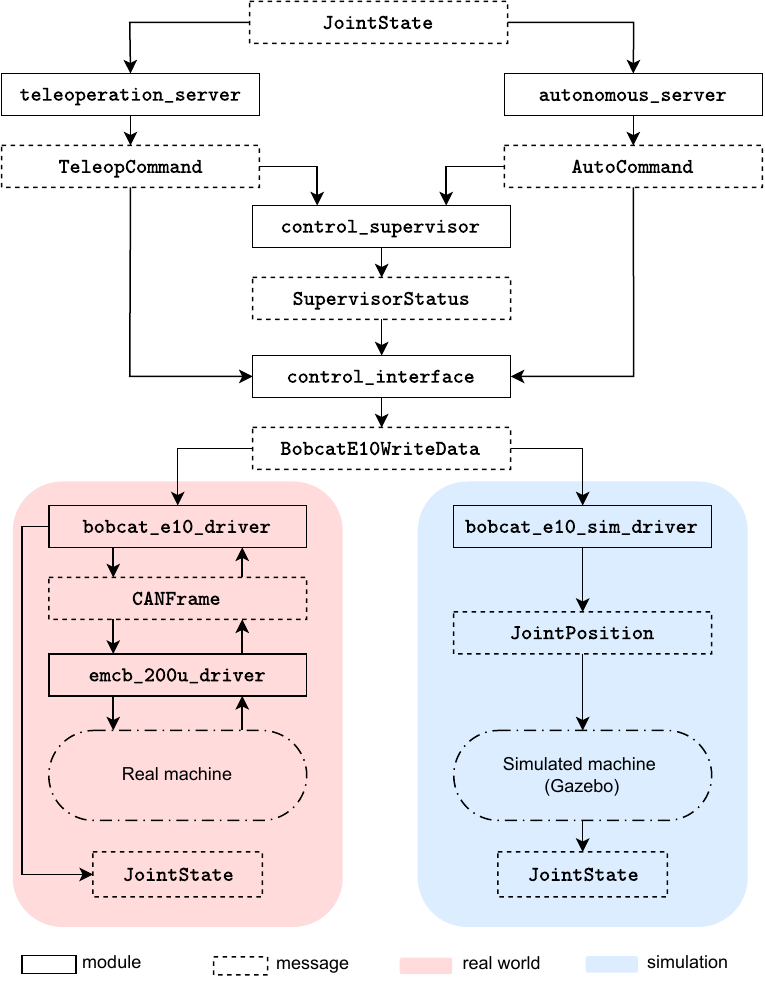}
    \caption{ROS-based pipeline for the reaching controller when deployed in simulation (using Gazebo) and in the real world.}
    \label{fig:bobcat_e10_ros_pipeline}
\end{figure}

To deploy reaching controllers in the real-world, a ROS-based pipeline was developed. This computational pipeline, along with its components and their interactions, is illustrated in Fig.~\ref{fig:bobcat_e10_ros_pipeline}. To minimize discrepancies between components, the pipeline only bifurcates from the drivers (which are distinct for the simulation in Gazebo and the real world) until the end, resulting in two distinct execution branches. 

For the pre-deployment stage, the learned dynamic models described in Section~\ref{subsec:dynamic_model_training_eval} were used to replicate the behavior of the real machine in the Gazebo simulator. To do so, these models are instantiated in the \texttt{bobcat\_e10\_sim\_driver} component. This module allows bypassing Gazebo’s built-in physics, enforcing the dynamics learned from the data obtained by operating the real machine, allowing the replication of hard-to-model effects, such as command-response delays. 

The reaching controller is encapsulated within the \texttt{autonomous\_server} module, which allows selecting among MPC and RL-based policies and provides them with the necessary data to populate the buffers they use to construct $\tilde{\bm{x}}_t$. This data is obtained from the \texttt{JointStates} messages, which specify the joint positions and velocities of the impact hammer. Through this server, the user can specify a target end-effector pose. The same server generates the corresponding commands, sends them to the subsequent components, maintains a fixed control frequency, and monitors terminal conditions to determine when to stop the hammer's operation, for instance, upon reaching the goal within a predefined tolerance, leaving the workspace, or exhibiting oscillatory behaviors. The two latter safety measures, however, are fully turned off during the experiments reported in this work, so as to measure the controllers' performance in an isolated manner.

At this stage, it can be observed that the commands generated by both the \texttt{teleoperation\_server} and the \texttt{autonomous\_server} follow the same pipeline. Whenever a command is sent, it is received by the \texttt{control\_supervisor} and the \texttt{control\_interface} modules. The \texttt{control\_supervisor} module is responsible for enforcing several safety constraints. It monitors factors such as control frequency, command delays, and potential self-collisions. If an abnormal condition is detected, it triggers an emergency stop signal to stop the hammer. This signal is received by the \texttt{control\_interface}, which converts the commands into a different message type that can be consumed by the simulated or real machine drivers. When the \texttt{control\_supervisor} issues an emergency signal, the \texttt{control\_interface} immediately mutes all outgoing commands. However, we must note that to isolate the performance of the real policies' during evaluations in Gazebo and in the real world, the \texttt{control\_supervisor} module remains disabled for all experiments conducted in this work.

Finally, in the case of the real machine, the \texttt{bobcat\_e10\_driver} translates the incoming ROS commands into CAN frames, which are received by the \texttt{emcb\_200u\_driver} and then written to the machine’s PLC. This component also reads information from the PLC and shares it with the Bobcat driver through CAN frames, which are parsed, pre-processed, and published to the rest of the system; in particular, this allows the population of data to construct the \texttt{JointState} messages of the real machine.

\section{Training details for dynamic models}
\label{appendix:dynamic_models_training_details}

Table~\ref{tab:optuna_optim} shows the search space of all the hyperparameters that are jointly optimized when attempting to learn dynamic models via different parameterizations (MLPs or KANs). Note that for a given parameterization and hyperparameter, the search space does not make a distinction between the type of prediction of the dynamic model ($\Delta\bm{q}$ or $\Delta\dot{\bm{q}}$). Also note that all hyperpamater search spaces are discrete and, except for $k$ (the number of lags used to construct the approximate state $\tilde{\bm{x}}_t$), are defined with values that are of common usage in standard machine learning practice. For $k$, the search space is defined considering the maximum measured actuation delay (around one second) when controlling each actuator of the mini-excavator arm with a constant joint command ($-1$ or $1$, prior to denormalization) at $20$ Hz.

When training MLPs, we use sigmoid linear unit (SiLU) activation functions for each hidden layer and a linear activation function for the output layer. When training KANs, we use Legendre basis functions for each layer. The instantiation of KANs in JAX is done using the jaxKAN package~\citep{rigas2025jaxkan}.
\begin{table}
    \caption{Search spaces for the dynamic models hyperparameters.}
    \label{tab:optuna_optim}
    \centering
    \begin{tabular}{cll}
    \toprule
     \textbf{Parameterization} &  \textbf{Hyperparameter}  & \textbf{Search space} \\ \midrule
     \multirow{3}{*}{MLP \& KAN}&  Nb. of lags $k$   &  $\{18, ..., 25\}$  \\
      &  Learning rate   &  $\{10^{-3}, 10^{-3}, 10^{-5}\}$  \\
     & Batch size & $\{256, 512, 1024, 2048\}$ \\ \midrule
     \multirow{3}{*}{MLP} &  Loss time horizon $H$   &  $\{4,8,16,32\}$  \\
     & Nb. of layers   &  $\{2,3,4,5\}$ \\
     &  Nb. of units per layer   &  $\{64, 128, 256, 512\}$ \\\midrule 
     \multirow{3}{*}{KAN} & Loss time horizon $H$   &  $\{8,16,32\}$  \\
     & Nb. of layers    & $\{3,4,5\}$ \\ 
      & Nb. of unites per layer     & $\{4,8,16,32\}$ \\ 
    \bottomrule
    \end{tabular}
    
\end{table}

To adjust the hyperparameters shown in Table~\ref{tab:optuna_optim}, we use the Optuna package. To cover the range of possibilities for the type of model parameterization (MLP or KAN) and prediction ($\Delta\bm{q}$ or $\Delta\dot{\bm{q}}$), four independent optimization processes are conducted, each of them for models parameterized using an MLP or a KAN, and predicting angular position or angular velocity residuals. The optimization, in all cases, is performed with respect to $\mathcal{L}(\bm{\theta})|_{H=80}$, which, as stated in Section~\ref{subsec:dynamic_model_training_eval}, corresponds to the loss function used for training with a fixed time horizon, $H=80$. 

In all cases, we train each model for a maximum of $1000$ and $1200$ epochs for MLP and KAN parameterizations, respectively, and use the validation subset to end a given trial if $\mathcal{L}(\bm{\theta})|_{H=80}$ does not decrease for $50$ consecutive training steps (early stopping). The optimization process allows us to obtain the four top dynamic models that are studied in Section~\ref{subsec:dynamic_model_training_eval}.

\section{Further details regarding policy synthesis}
\label{appendix:policy_learning_details}

Table~\ref{tab:modeling_algo_hypeparams} shows the hyperparameters and the values they take for the control and modeling aspects associated with the reaching task. This table also presents the parameters set for policy synthesis using PPO and iCEM.

\begin{table}
    \caption{Modeling and algorithms hyperparameters.}
    \label{tab:modeling_algo_hypeparams}
    \centering
    \begin{tabular*}{\linewidth}{l@{\extracolsep{\fill}}l@{\extracolsep{\fill}}l@{\extracolsep{\fill}}}
    \toprule
        & \textbf{Parameter} & \textbf{Value} \\ \midrule
         \multirow{3}{*}{Control}  & Control frequency [Hz] & $20$\\ 
         & Action repeats & $4$ \\ 
         & Loop tracker $k_p, k_i$ & $10.0, 60.0$ \\ \midrule
        \multirow{3}{*}{\begin{tabular}[c]{@{}l@{}}Episodic\\ conditions\end{tabular}}   & $\alpha^\text{max}_\text{pitch}, \alpha^\text{ts}_\text{pitch}$ [rad] & $1.047, 0.698$\\ 
        & $z^{\text{ts}}_\text{min}, z^{\text{ts}}_\text{max}$ [m]& $0.155, 2.355$ \\
        & Max number of steps $T_\text{reset}$ & 500 \\ \midrule
       \multirow{3}{*}{\begin{tabular}[c]{@{}l@{}}Reward\\ function\end{tabular}} & $\lambda_{\bm{p}}, \lambda_{\bm{r}}, \lambda_{\bm{q}}, \lambda_{\bm{a}}, \lambda_{\mathcal{W}^+}$ & $1.0, 1.0, 1.0, 0.5, 2.0$\\
       & $\epsilon_{\bm{p}}, \epsilon_{\bm{r}}, \epsilon'_{\bm{p}}, \epsilon'_{\bm{r}}$&  $0.1, 0.1, 0.05, 0.02$\\
       & $\epsilon_{\alpha}, \epsilon_q, w_\mathcal{X}, w_\alpha, w_{\bm{q}}$& $0.0001, 0.0025, 0.5, 1.0, 1.0$ \\
       \midrule 
       \multirow{12}{*}{PPO} & Training steps & $400$M \\
           & Learning rate & $3\cdot10^{-4}$\\ 
           & Entropy cost & $10^{-2}$\\
           & Discount factor $\gamma$ & $0.97$ \\ 
           & Unroll length & $40$\\ 
           & Clipping $\epsilon$ & 0.2 \\
           & Batch size & $256$\\
           & Number of mini-batches & $32$\\
           & Number of environments & $128$\\
           & Updates per batch & $4$\\
           & Policy parameterization & $\text{MLP}(256,)_{\times4}$\\
           & Critic parameterization & $\text{MLP}(256,)_{\times 5}$\\
           \midrule
       \multirow{9}{*}{iCEM} 
           & Horizon $H_{\text{MPC}}$ & $20$ \\
           & Number of samples $N_\text{pop}$ & $500$ \\
           & Number of elites $N_e$ & $50$ \\
           & Initial std $\sigma_0$ & $0.5$ \\
           & Smoothing coefficient $\alpha$ & $0.05$ \\
           & Optimization steps $K$ & $12$ \\
           & Exponent & $2.0$ \\
           & Elite set fraction & $0.5$ \\
           & Action repeats & $4$ \\
    \bottomrule
    \end{tabular*}
    
\end{table}

The policy learning process using PPO is conducted using a computer equipped with an Intel i7-12700 CP CPU, a NVIDIA RTX 4060 GPU, and 32 GB of RAM. Training the reaching policies for $400\text{M}$ of steps takes approximately $6$ min. on average. The iCEM controller instantiated with the parameters reported in Table~\ref{tab:modeling_algo_hypeparams} (running on the same computer) averages an optimization computation time of approx. $90$ ms. This result makes this method suitable for controlling the machine at $20$~Hz when using $4$ action repeats, considering that the experiments conducted in Gazebo (Sim2Sim) show that the lag in the generation of control commands that the above induces does not affect performance severely (see Section~\ref{subsec:sim2sim}). 

\begin{table*}[]
    \sisetup{separate-uncertainty=true, table-align-uncertainty=true}
    \centering
    \caption{Performance of the PPO MLP $\Delta \dot{\bm{q}}$ policy when trained using different reward functions.}
    \label{tab:reward_ablation}
    \begin{tabular*}{\linewidth}{l@{\extracolsep{\fill}}S[table-format=-3.2(5)]@{\extracolsep{\fill}}@{\extracolsep{\fill}}c@{\extracolsep{\fill}}c@{\extracolsep{\fill}}c@{\extracolsep{\fill}}c@{\extracolsep{\fill}}c@{\extracolsep{\fill}}c}
        \toprule
        \textbf{Reward function}  & {\textbf{Avg. Return}$^{*}$} &\textbf{SR}$\bm{(0.02, 0.02)}$ & \textbf{SR}$\bm{(0.04, 0.04)}$ & \textbf{OO}$\bm{\mathcal{W}}$ & \textbf{OO}$\bm{Z}$ & \textbf{Min.} $\bm{z}$ \textbf{[m]} & \textbf{EEF-PL [m]}\\ \midrule
         $r_t^{\mathcal{X}} $ & -201.65 \pm 98.81 & $0.30\pm0.33$&  $0.44\pm0.38$& $0.22\pm0.04 $ & $0.12\pm0.03$ & $-0.20\pm0.12$ & $4.50 \pm1.04$\\
         $r_t^{\bm{q}}$ & -91.17 \pm 0.49& $0.97 \pm 0.02$ & $1.00 \pm 0.00$ & $0.21 \pm 0.01$ & $0.16 \pm 0.01$ & $-0.16 \pm 0.01$ & $2.41 \pm 0.04$\\
         $r_t^{\mathcal{X}} + r_t^\epsilon$ & 842.83 \pm 20.26 & $1.00 \pm 0.01$ &  $1.00 \pm 0.00$& $0.16 \pm 0.01$ & $0.12 \pm 0.01$&$-0.24 \pm 0.06$  & $2.12 \pm 0.04$ \\
         $r_t^{\bm{q}} + r_t^\epsilon$ & 870.74 \pm 10.72  &$1.00 \pm 0.00$ & $1.00 \pm 0.00$  & $0.22 \pm 0.01$ & $0.17 \pm 0.01$ &  $-0.19 \pm 0.09$ & $2.10 \pm 0.02$ \\
         $r_t^{\mathcal{X}} + r_t^{\bm{q}}$ & -290.18 \pm 68.68 & $0.59 \pm 0.35$ & $0.77 \pm 0.21$& $0.27 \pm 0.10$& $ 0.20 \pm 0.09$& $-0.30\pm 0.21$ &$3.83 \pm 1.37$\\
         $r_t^{\mathcal{X}} + r_t^{\bm{q}} + r_t^{\epsilon} $ & 753.17 \pm 12.60 & $1.00\pm 0.00$ & $1.00\pm0.00$ & $0.18\pm 0.01$& $0.13\pm 0.01$& $-0.24\pm0.06$ & $2.10\pm0.03$\\
         $r_t^{\mathcal{X}} + r_t^{\bm{q}} + r_t^{\epsilon} + r_t^{\bm{a}}$ & 719.32 \pm 14.33 & $0.99\pm 0.01$ & $1.00 \pm 0.00$& $0.19\pm 0.02$&$0.14\pm 0.01$ & $-0.24\pm 0.06$ & $2.14\pm0.03$\\
         $r_t^{\mathcal{X}} + r_t^{\bm{q}} +  r_t^{\bm{a}} + r_t^{\mathcal{W}}$ & -356.61 \pm 203.17& $0.59 \pm 0.30$& $0.78 \pm 0.37$& $0.16 \pm 0.04$ &$0.10 \pm 0.03$ & $-0.12 \pm 0.11$ & $3.97 \pm 1.77$\\
         $r_t^{\mathcal{X}} + r_t^{\bm{q}} + r_t^{\epsilon} + r_t^{\bm{a}} + r_t^{\mathcal{W}}$  & 683.36 \pm 16.94& $0.99 \pm 0.01$& $1.00 \pm 0.00$&  $0.15 \pm 0.01$ & $0.11 \pm 0.01$ & $-0.10 \pm 0.08$ & $2.15 \pm 0.05$ \\
        \bottomrule
    \end{tabular*}
    \vspace{2pt}
    
    \footnotesize{\raggedright$^{*}$ For each row, the computation of returns depends on the reward function used for training.\hfill}
    
\end{table*}

For the PPO policy trained using the learned MLP~$\Delta \dot{\bm{q}}$ model, an ablation study on the components of the reward function is provided in Table~\ref{tab:reward_ablation}. Besides the evaluation metrics used in the results presented in Sections~\ref{subsec:training_eval_brax} and~\ref{subsec:sim2sim_andsim2real}, we include the avg. traversed end-effector path length (EEF-PL), which is computed by summing the Euclidean distance between successive end-effector positions for a given episode up until the target is reached with a certain precision ($0.02$~[cm] and $0.02$~[rad]), or until $T_{\text{max}}$ steps pass. The distances obtained are then averaged across the evaluation episodes.

The results suggest that including the $r_t^{\epsilon}$ term in the reward function definition is crucial for achieving high success rates at high precision for controllers trained using PPO. 

The results also show that good performance in terms of success rate can be achieved by relying on $r_t^{\epsilon}$ alongside either $r_t^{\bm{q}}$, $r_t^{\mathcal{X}}$ or both terms. We must note that this is expected, since both $r_t^{\bm{q}}$ and $r_t^{\mathcal{X}}$ serve the purpose of guiding the robot towards the target pose; however, combining $r_t^{\bm{q}}$ and $r_t^{\mathcal{X}}$ without the bonus term results in high variance across training trials. We attribute this behavior to $r_t^{\bm{q}}$ and $r_t^{\mathcal{X}}$ being potentially conflicting, and $r_t^{\epsilon}$ serving as a term that dominates the optimization landscape due to the high reward the agent receives if it reaches the target pose. Finally, we must also note that, regardless of the potential tension between terms, we choose to use both $r_t^{\bm{q}}$ and $r_t^{\mathcal{X}}$ in the final reward function definition because the agent's observations are constructed using both position-quaternion tuples and the associated configurations to represent current and target end-effector poses.

\section{Further results regarding the Sim2Sim transfer experiments}
\label{appendix:sim2sim}

\begin{figure*}
    \centering
    \subfloat[][]{\includegraphics[width=\linewidth]{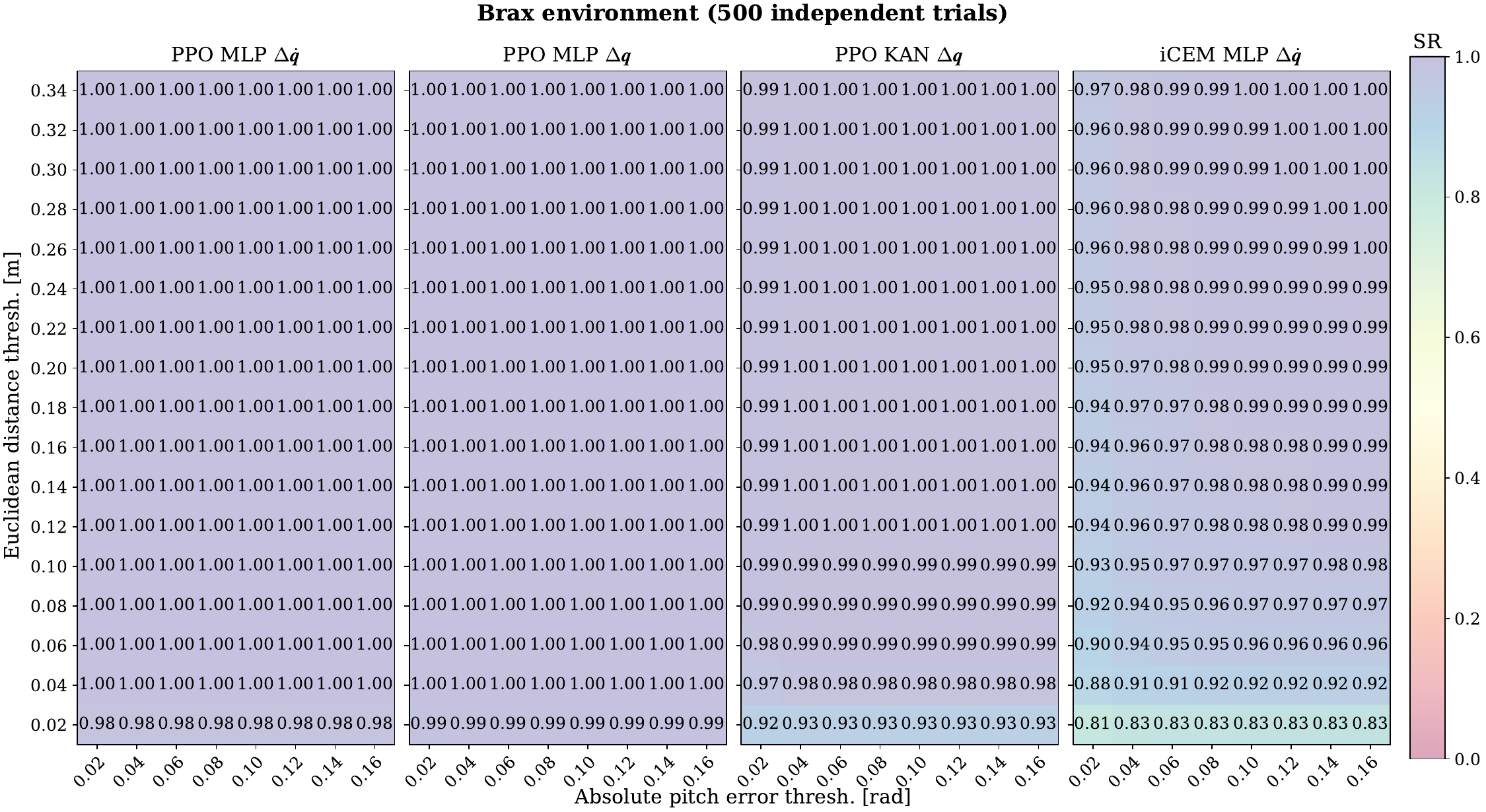}\label{fig:sr_matrices_brax}}\\
    \subfloat[][]{\includegraphics[width=\linewidth]{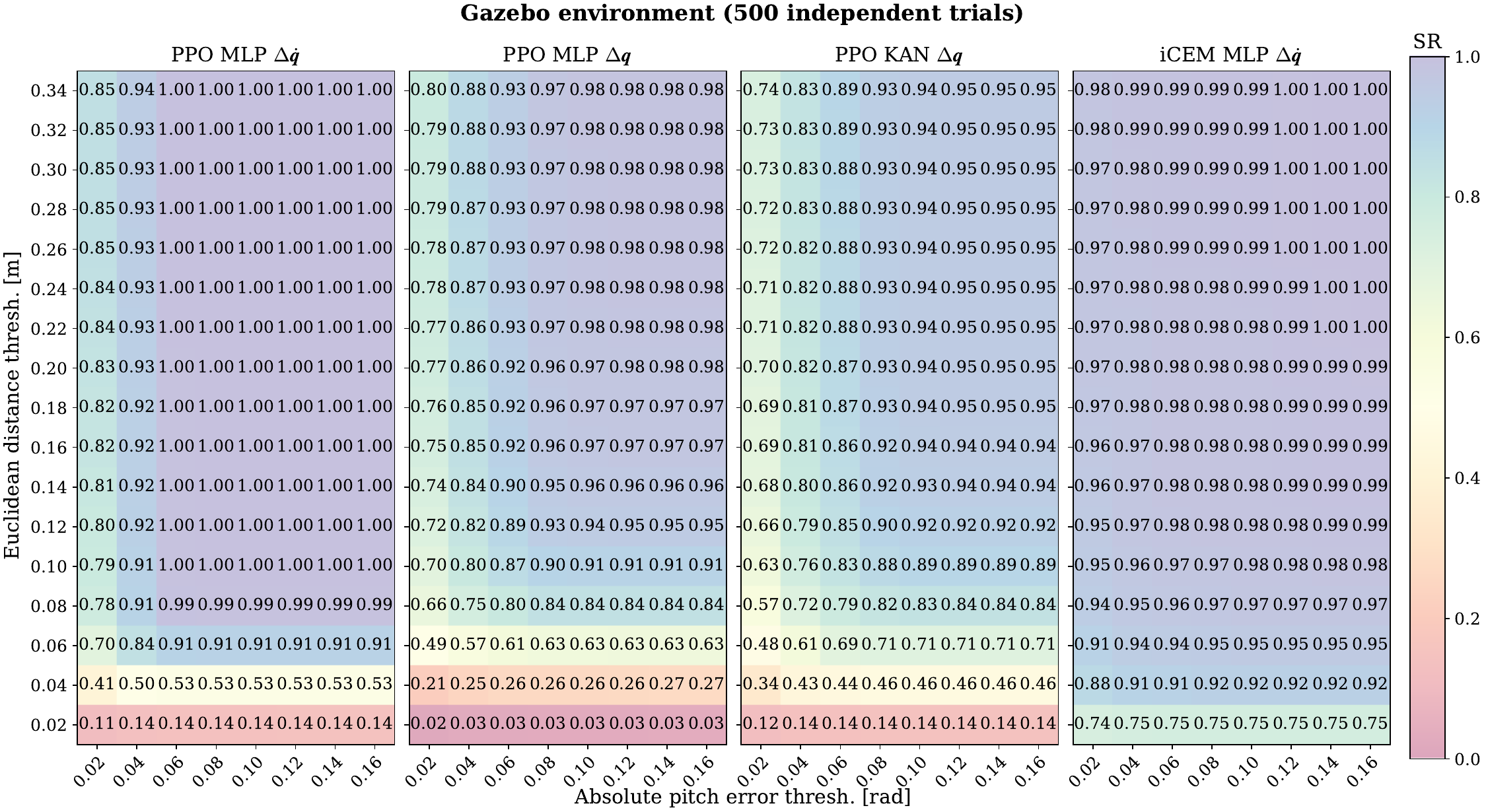}\label{fig:sr_matrices_gazebo}}
    \caption{Success rate matrices obtained when evaluating the reaching policies in (a) the Brax-based environment used for training, and (b) the Gazebo simulator, by encapsulating the policies in action servers.}
    \label{fig:sr_matrices_brax_gazebo}
\end{figure*}

Fig.~\ref{fig:sr_matrices_brax_gazebo} shows the success rate matrices obtained by evaluating the RL- and MPC-based reaching policies in the Brax-based environment in which they were synthesized (Fig.~\ref{fig:sr_matrices_brax}) and the matrices obtained when evaluating the same policies in Gazebo (Fig.~\ref{fig:sr_matrices_gazebo}). For a detailed analysis on these results, refer to Section~\ref{subsec:sim2sim}.

\section*{Acknowledgments}

The authors thank Claudio Palacios for designing and implementing the electro-hydraulic interventions that were required to automate the Bobcat~E10. The authors also acknowledge Pablo Alfessi for creating a CAD model of the Bobcat~E10, and for designing, manufacturing, and installing the mountings of the rotary's encoders on the robot's arm. Finally, the authors also thank Cristian Rivera for implementing most of the additional instrumentation required for automating the machine, and for designing and implementing the program that runs on the PLC.

\balance
\bibliographystyle{apalike}
\bibliography{references}

\end{document}